\documentclass[11pt]{article}
\usepackage{algorithm}
\usepackage{algpseudocode}
\usepackage{amsfonts}
\usepackage{amsmath}
\usepackage{amssymb}
\usepackage{authblk}
\usepackage{booktabs}
\usepackage[margin=1in]{geometry}
\usepackage{graphicx}
\usepackage{multirow}
\usepackage{xcolor}
\usepackage[colorlinks=true,citecolor=blue]{hyperref}

\algrenewcommand\algorithmicrequire{\textbf{Input:}}
\algrenewcommand\algorithmicensure{\textbf{Output:}}
\newcommand{\AlgBlockComment}[1]{%
  \Statex \vspace{0.8em}\textit{#1}%
}

\title{PI-MFM: Physics-informed multimodal foundation model for solving partial differential equations}
\author[1]{Min Zhu}
\author[2]{Jingmin Sun}
\author[3]{Zecheng Zhang}
\author[4]{Hayden Schaeffer}
\author[1,5,*]{Lu Lu}
\affil[1]{Department of Statistics and Data Science, Yale University, New Haven, CT 06511, USA}
\affil[2]{Department of Applied Mathematics and Statistics, Johns Hopkins University, Baltimore, MD 21218, USA}
\affil[3]{Department of Applied Computational Mathematics and Statistics, University of Notre Dame, Notre Dame, IN 46556, USA}
\affil[4]{Department of Mathematics, University of California Los Angeles, Los Angeles, CA 90095, USA}
\affil[5]{Department of Chemical and Environmental Engineering, Yale University, New Haven, CT 06511, USA}
\affil[*]{Corresponding author. Email: lu.lu@yale.edu}

\date{}

\begin{document}

\maketitle

\begin{abstract}
Partial differential equations (PDEs) govern a wide range of physical systems, and recent multimodal foundation models have shown promise for learning PDE solution operators across diverse equation families. However, existing multi-operator learning approaches are data-hungry and neglect physics during training. Here, we propose a physics-informed multimodal foundation model (PI-MFM) framework that directly enforces governing equations during pretraining and adaptation. PI-MFM takes symbolic representations of PDEs as the input, and automatically assembles PDE residual losses from the input expression via a vectorized derivative computation. These designs enable any PDE-encoding multimodal foundation model to be trained or adapted with unified physics-informed objectives across equation families. On a benchmark of 13 parametric one-dimensional time-dependent PDE families, PI-MFM consistently outperforms purely data-driven counterparts, especially with sparse labeled spatiotemporal points, partially observed time domains, or few labeled function pairs. Physics losses further improve robustness against noise, and simple strategies such as resampling collocation points substantially improve accuracy. We also analyze the accuracy, precision, and computational cost of automatic differentiation and finite differences for derivative computation within PI-MFM. Finally, we demonstrate zero-shot physics-informed fine-tuning to unseen PDE families: starting from a physics-informed pretrained model, adapting using only PDE residuals and initial/boundary conditions, without any labeled solution data, rapidly reduces test errors to around 1\% and clearly outperforms physics-only training from scratch. These results show that PI-MFM provides a practical and scalable path toward data-efficient, transferable PDE solvers.
\end{abstract}


\section{Introduction}

Partial differential equations (PDEs) model a wide range of phenomena by relating physical quantities to their rates of change. PDEs appear across fluid dynamics, structural mechanics, seismology, and many other areas, offering a mathematical basis for analyzing and predicting complex systems. Physics-informed neural networks (PINNs)~\cite{raissi2019physics,karniadakis2021physics,lu2021deepxde} combine deep learning with governing equations by embedding the PDE into the training objective. This allows solving forward and inverse PDE problems on continuous domains without predefined meshes, in contrast to classical finite-difference and finite-element solvers that require discretization. PINNs are effective at jointly leveraging real-world data and physical constraints, but they typically need to be retrained whenever the PDE, initial condition (IC), or boundary condition (BC) changes, which is costly for problems that change frequently.

Operator learning addresses this challenge by learning PDE-induced solution operators, i.e., mappings between function spaces from input fields to the corresponding solutions. Deep operator network (DeepONet)~\cite{lu2021learning} and its variants~\cite{lu2022multifidelity,zhu2023reliable,zhu2023fourier,jiao2024solving,jiang2024fourier,lee2024efficient,yin2024scalable,xiao2025quantum,kou2025neural,jiao2025one,ouyang2025rams}, as well as Fourier neural operator (FNO)~\cite{li2020fourier} and its modifications~\cite{wen2022u,wen2023real,zhu2023fourier,jiang2024fourier,lee2024efficient}, learn solution operators that can adapt to varying inputs more easily than instance-specific PINNs. Recent work has also explored operator learning using diffusion models~\cite{wang2025fundiff,wang2025geofunflow,shan2025red} and transformer-based architectures~\cite{wu2025tante,liu2024prose,sun2025towards,sun2024lemon}. Early operator-learning methods were purely data-driven and often required large labeled datasets. Physics-informed variants, such as physics-informed DeepONet~\cite{wang2021learning} and PINO~\cite{li2024physics}, incorporate governing equations into training, improving data efficiency and physical fidelity. Most of these methods, however, were developed in the single-operator learning (SOL) regime, which targets one operator at a time.

Recent work has shifted from single-operator learning to multi-operator learning (MOL)~\cite{jin2022mionet}, where a single neural network is trained over a collection of PDEs. Many recent MOL methods follow the multimodal foundation model paradigm~\cite{bommasani2021opportunities}: they use large networks pretrained on diverse PDE families and ingest multiple modalities such as coefficient and forcing fields, boundary and initial conditions, and sometimes explicit PDE descriptors~\cite{liu2024prose,sun2025towards,sun2024lemon,mccabe2024multiple,hao2024dpot}. These MOL approaches can be broadly divided into two classes: non-PDE-encoding methods and PDE-encoding methods, depending on whether an explicit representation of the PDE (e.g., symbolic tokens or an equation graph) is provided as a network input modality. Non-PDE-encoding methods do not supply such an explicit PDE encoding; instead, they condition on observed function data without explicitly identifying the underlying PDE. Representative examples include the in-context operator network (ICON)~\cite{yang2023context,yang2024pde,yang2025fine} and transformer-based PDE foundation models~\cite{mccabe2024multiple,hao2024dpot,shen2024ups,herde2024poseidon,liu2025bcat}. PDE-encoding methods, in contrast, feed an explicit representation of the governing PDE into the network, for example symbolic PDE expressions in PROSE~\cite{liu2024prose,sun2025towards,sun2024lemon}, embeddings of PDE components (equation symbols, coefficients, and conditions) in Unisolver~\cite{zhou2024unisolver}, and graph-structured encodings of PDEs in PDEformer~\cite{ye2024pdeformer,ye2025pdeformer}. 

Despite these advances, existing MOL systems often require substantial labeled data and tend to treat physics (e.g., PDEs) as side information rather than as constraints in the training objective. In this work, we introduce physics-informed multimodal foundation models (PI-MFM), an PDE-encoding MOL framework that directly enforces physics during both pretraining and adaptation. Our work adopts PDE-encoding multimodal foundation models since we aim for the model to directly learn physical information. Concretely, PI-MFM takes as input both the ICs/BCs and symbolic representation of the PDE, and is trained by combining physics losses with only a small number of labeled data. This design significantly improves data efficiency and generalization across diverse PDE families while retaining the flexibility of MOL.

\paragraph{Implementation and generality.}
In this paper, we use PROSE~\cite{liu2024prose,sun2024lemon,sun2025towards} as the multimodal foundation model backbone in the proposed PI-MFM framework. PROSE provides an PDE-encoding interface that accepts PDE information in the form of symbolic expressions. Based on these symbolic inputs, our framework automatically constructs and injects physics losses (PDE residuals and IC/BC terms) during training and adaptation. Importantly, the physics-informed formulation does not depend on the network architecture: any multimodal foundation model that accepts PDE information (e.g., symbolic tokens or structured PDE descriptors) can be trained or adapted with the PI-MFM framework. In addition, given a symbolic PDE expression as input, our framework vectorizes the computation of all required derivatives at collocation points and automatically assembles the corresponding physics loss, obviating manual derivations and custom PDE-specific loss implementations.

\paragraph{Comparison with prior physics-informed methods.}
A practical difference is that prior physics-informed methods, such as PINN, PINO~\cite{li2024physics}, and physics-informed DeepONet~\cite{wang2021learning}, require hand-crafted physics loss terms for each PDE~\cite{zhang2024deeponet}. In contrast, our PI-MFM framework computes these losses automatically from the input PDE expression and conditions.
Moreover, prior physics-informed methods primarily target SOL, or they do not explicitly encode the PDE as a symbolic input to a single multi-operator network. Our setting is different: we explicitly encode the symbolic PDE expression and train one model over multiple PDE families. Because existing physics-informed baselines are not designed for this explicit PDE-encoding MOL regime, there is no directly comparable baseline. Therefore, throughout the paper, we compare the models trained with physics losses (``w/ physics'') and without physics losses (``w/o physics'') under consistent data, sampling, and optimization setups to demonstrate the benefits of our approach.

\paragraph{Benefits of PI-MFM.}
Across our experiments in Section~\ref{sec:results}, we observe consistent gains from the PI-MFM framework.
\begin{itemize}
    \item First, it is data-efficient: with sparse labeled spatial and temporal data or function data, the ``w/ physics'' models achieve much lower relative errors than purely data-driven counterparts under the same supervision budgets.
    \item Second, it is robust: under large label noise, physics losses provide an effective physics-guided prior and regularizer, mitigating overfitting when labels are scarce or corrupted.
    \item Third, simple training strategies, such as resampling collocation points for the PDE residual, further reduce error without changing the architecture.
    \item Fourth, it is practical: our vectorized PDE loss computation algorithm evaluates all required coordinate derivatives in a single batched pass. Moreover, our analysis and experiments on automatic differentiation versus finite differences provide feasible guidance for selecting an appropriate differentiation backend when implementing PDE residual losses in PDE-encoding foundation models.
\end{itemize}

Remarkably, PI-MFM enables zero-shot physics-informed fine-tuning to unseen PDE families. Starting from a physics-informed pretrained model, we adapt to a new PDE family using only physics supervision, i.e., PDE residuals and IC/BCs, without any labeled input/output pairs, yet rapidly attain about $1\%$ relative error and outperform training-from-scratch models that use the same physics-only losses. This demonstrates that physics-informed multimodal foundation models can transfer across PDE families using purely physics-based supervision, providing a practical approach to fast, accurate adaptation in new scientific regimes.

\label{sec:intro}
\section{Problem setup and datasets}
\label{sec:setup}

In this section, we describe the problem setting and datasets used in our study.
In Section~\ref{subsec:pdes_define}, we introduce the families of parametric PDEs considered in our experiments, together with their initial and boundary conditions.
In Section~\ref{subsec:polish}, we explain how PDEs are encoded as symbolic expressions using Polish notation.
Finally, in Section~\ref{subsec:sol_mol}, we distinguish between single-operator learning (SOL) and multi-operator learning (MOL), and formulate the general learning task addressed in this work.

\subsection{PDE families}
\label{subsec:pdes_define}

% PDEs whose equations do not explicitly depend on the independent variables $t$ and $x$, but only on the dependent variable $u$ (autonomous)

Let $\mathcal{I}$ denote an index set labeling different families of PDEs. Each PDE family $i \in \mathcal{I}$ is associated with a parameter vector $ \mathbf{q}^i \in \mathcal{D}^i $, where $\mathcal{D}^i \subseteq \mathbb{R}^{m_i}$ is the parameter space of the $i$-th PDE family, and $m_i$ is the dimension of parameters $\mathbf{q}^i$. Given $i \in \mathcal{I}$ and $\mathbf{q}^i \in \mathcal{D}^i$, the parametric PDE is written as
% We equip each PDE family with parameters $ \mathbf{q}^i \in \mathcal{D}^i $, where $\mathcal{D}^i \subseteq \mathbb{R}^{m_i}$ is the parameter space of the PDE family $i$, and $m_i$ is the dimension of parameters $\mathbf{q}^i$. 
\begin{equation}
\label{eq:pde}
    \mathcal{F} ( u; i, \mathbf{q}^i ) = 0,
\end{equation}
supplemented with proper initial conditions (ICs) and boundary conditions (BCs).

% By definition, these autonomous PDEs do not explicitly depend on the independent variables $t$ and $x$; rather, they involve only the dependent variable $u$.

% The autonomous PDEs do not explicitly contain the independent variables, $t$ and $x$, but only involve the dependent variable $u$.

In this work, we consider one-dimensional, time-dependent PDEs defined for the unknown function 
$$u: [0,1] \times [0,1] \to \mathbb{R}, \quad (t,x) \mapsto u(t,x),$$
subject to periodic boundary condition and initial condition
$$u(0,x)=u_0(x),$$
where $u_0(x)$ is constructed as a superposition of sinusoidal waves~\cite{takamoto2022pdebench}:
\begin{equation}
    \label{eq:ic}
    u_0(x) = \sum_{j=1}^{N} A_j \sin(k_j x + \phi_j)    
\end{equation}
with $ k_j = 2\pi n_j/{L_x} $ denoting the wavenumber. Here, each $ n_j $ is an integer randomly selected from the range $[1, n_{\max}]$. $ N $ specifies the number of waves, and $ L_x =1$ is the length of the spatial domain. The amplitude $ A_j $ is a random real number sampled uniformly from $[0,1]$, and the phase $ \phi_j $ is chosen uniformly from $[0, 2\pi)$. In the work, we set $n_{\max}=4$ and $N=2$. The initial condition values are rescaled to $[0,1]$ via Min-Max normalization.
Let $\mathcal{I}^\prime \subseteq \mathcal{I}$ denote the subset of indices corresponding to second-order-in-time PDEs. For these PDEs, an additional initial condition IC$^\prime$ is 
$$
\frac{\partial u}{\partial t} (0, x) = 0.
$$

%In this work, we consider 1st and 2nd order-in-time PDEs. For initial condition $\frac{\partial u}{\partial t}(0,x) = u_1(x)$, we consider the case $\frac{\partial u}{\partial t}(0,x) = 0$. 

The PDEs considered in this work are classified into 13 families, each characterized by variable parameters (Table~\ref{tab:pdes}). For the $i$-th family, the parameter vector $\mathbf{q}^i$ may include either a single component, $q$, or two components, $q$ and $p$. The central values of these parameters are provided in the ``Parameter center'' column of Table~\ref{tab:pdes}. During data generation, $q$ and $p$ are randomly sampled from the intervals $[90\% q_c, 110\% q_c]$ and $[90\% p_c, 110\% p_c]$, respectively, where $q_c$ and $p_c$ denote the parameter centers. In the ``Equation'' column of Table~\ref{tab:pdes}, $u_t$ and $u_x$ denote the first-order partial derivatives of the solution $u$ with respect to $t$ and $x$, respectively, while $u_{tt}$ and $u_{xx}$ denote the corresponding second-order derivatives.

\begin{table}[htbp]
    \centering
    \small
    \caption{\textbf{13 PDE families with their equations and parameters.} Parameters $q$ and $p$ are randomly sampled from the intervals $[90\% q_c, 110\% q_c]$ and $[90\% p_c, 110\% p_c]$, respectively, where $q_c$ and $p_c$ denote the parameter centers.}
    \begin{tabular}{c|c|l|l}
        \toprule
        PDE family & Abbreviation & Equation & Parameter center \\
        \midrule
        Advection & Adv & $u_t =- q u_x$ & $q_c = 0.5$ \\
        Diffusion & Diff & $u_t = q u_{xx}$ & $q_c = 0.003$ \\
        Diffusion Linear Reaction & Diff-Lin & $u_t = q u_{xx} + p u$ & $q_c = 0.003, p_c = 0.1$ \\
        Diffusion Logistic Reaction & Diff-Log & $u_t = q u_{xx} + p u (1 - u)$ & $q_c = 0.003, p_c = 1$ \\
        Diffusion Square Logistic Reaction & Diff-SLog & $u_t = q u_{xx} + p u^2 (1 - u)^2$ & $q_c = 0.003, p_c = 1$ \\
        Diffusion Bistable Reaction & Diff-Bi & $u_t = q u_{xx} + p u^2 (1 - u)$ & $q_c = 0.003, p_c = 1$ \\
        Conservation Law with Cubic Flux & Cons-Cub & $u_t = -q \left( \frac{u^3}{3} \right)_x + \frac{p}{\pi} u_{xx}$ & $q_c = 1, p_c = 0.01$ \\
        Conservation Law with Linear Flux & Cons-Lin & $u_t = -q (u)_x + \frac{p}{\pi} u_{xx}$ & $q_c = 1, p_c = 0.01$ \\
        Conservation Law with Sine Flux & Cons-Sin & $u_t = -q (\sin u)_x + \frac{p}{\pi} u_{xx}$ & $q_c = 1, p_c = 0.01$ \\
        Burgers' & Burgers' & $u_t = -q \left( \frac{u^2}{2} \right)_x + \frac{p}{\pi} u_{xx}$ & $q_c = 1, p_c = 0.01$ \\
        Wave & Wave & $u_{tt} = q^2 u_{xx}$ & $q_c = 0.5$ \\
        Klein-Gordon & KG & $u_{tt} = -p^2 q^4 u + q^2 u_{xx}$ & $q_c = 1, p_c = 0.1$ \\
        Sine-Gordon & SG & $u_{tt} = -q \sin(u) + u_{xx}$ & $q_c = 1$ \\
        \bottomrule
    \end{tabular}
    \label{tab:pdes}
\end{table}

Of the 13 PDE families, 10 families (Adv, Diff, Diff-Lin, Diff-Log, Diff-SLog, Cons-Cub, Cons-Lin, Cons-Sin, KG, and SG) are used for training and testing our physics-informed foundation models. The remaining 3 PDE families (Diff-Bi, Burgers, and Wave) are reserved for downstream tasks to evaluate the generalization ability of our models.

\subsection{Polish notation and symbolic expression}
\label{subsec:polish}

A key aspect of multi-operator learning is the representation that enables models to capture both shared structure and operator-specific differences across PDE families. Following prior work on symbolic regression and operator learning~\cite{arabshahi2018combining,lample2019deep}, we represent each PDE as a symbolic expression composed of mathematical operators and functions acting on variables, derivatives, and constants. Such an expression naturally corresponds to a tree, where internal nodes are operators and leaves are operands (Fig.~\ref{fig:architecture}B). In our implementation, each PDE operator is serialized as a token sequence in Polish (prefix) notation~\cite{pogorzelski1965jan}, in which operators precede their operands, yielding a compact and unambiguous format that can be consumed directly by the symbol encoder.

When constructing physics-informed losses, we parse each token sequence once into an expression tree, and then recursively evaluate the tree using the network predicted solution and derivative channels to obtain discrete PDE residuals (Algorithm~\ref{alg:pde_loss_batch}). For additional details on symbolic encodings of PDE operators, see~\cite{liu2024prose,sun2025towards,d2022deep,kamienny2022end,tai2015improved}. For example, the advection equation $u_t + 0.514\,u_x = 0$ is encoded as \([\texttt{add}, u_t, \texttt{mul}, \texttt{add}, \texttt{N514}, \texttt{E-3}, u_x]\), where \([\texttt{add}, \texttt{N514}, \texttt{E-3}]\) encodes the parameter value $0.514$.

Let $s(i,\mathbf{q}^i)$ denote the symbolic expression represented in Polish notation of a PDE with family index $i$ and parameter $\mathbf{q}^i$, and $\mathcal{S}$ denotes the space of all possible symbolic expressions. Since the symbolic expression $s$ uniquely specifies both the PDE family and its parameters, we may redefine the PDE operator $\mathcal{F}$ to take $s$ as an argument directly. The governing equation, previously given in Eq~\eqref{eq:pde}, is now more generally expressed as
\begin{equation*}
\label{eq:pde_new}
    \mathcal{F} (u; s) = 0.
\end{equation*}
% In this form, the functional $\mathcal{F}$ acts as an interpreter, applying the differential operators specified by $s$ to the solution $u$.

\subsection{Single-operator learning and multi-operator learning}
\label{subsec:sol_mol}

In this work, we aim to learn the mapping from the initial condition $u_0(x)$ to the corresponding solution $u(t,x)$ for the PDEs defined in Section~\ref{subsec:pdes_define}.

\paragraph{Single-operator learning (SOL).}
Given a symbolic expression $s$ corresponding to a particular PDE, we define the associated solution operator $\mathcal{G}_{s}$ as
$$
  \mathcal{G}_{s} : \mathcal{U}_0 \to \mathcal{U},
  \quad
  u_0 \mapsto u,
$$
where $\mathcal{U}_0$ is the space of initial conditions $u_0(x)$ generated by Eq.~\eqref{eq:ic}, and $\mathcal{U}$ denotes the space of solutions. In the SOL setting, a model is trained specifically to approximate $\mathcal{G}_{s}$ for a fixed $s$. However, if the symbolic expression $s$ changes, the trained SOL model does not generalize and need to be retrained for the new PDE.

\paragraph{Multi-operator learning (MOL).}
In contrast, multi-operator learning (MOL) aims to learn a single model that approximates the family of solution operators $\bigl\{ \mathcal{G}_{s} \mid s \in \mathcal{S} \bigr\}$ for a collection of PDEs. Specifically, MOL addresses a more general task of learning the mapping
$$
\mathcal{G}: \mathcal{U}_0 \times \mathcal{S} \to \mathcal{U}, \quad (u_0, s) \mapsto u.
$$
By leveraging this symbolic encoding, MOL enables generalization across a wide variety of PDEs without the need for retraining, offering significantly greater flexibility compared to SOL methods.
To approximate this multi-operator mapping $\mathcal{G}$, we train a surrogate model $\mathcal{G}_\theta$, parameterized by $\theta$: 
\[
\hat u_\theta := \mathcal{G}_\theta(u_0,s), \qquad \hat u_\theta \approx u,
\]
where $\hat u_\theta $ is the predicted solution corresponding to the initial condition $ u_0 $ and symbolic PDE representation $s$. This framework allows the surrogate model $\mathcal{G}_\theta$ to generalize effectively across various PDE families and parameter settings by explicitly incorporating symbolic expressions as inputs.

% To address the limited extrapolation capability of SOL, we utilize a multi-operator learning (MOL) framework. Define
% $$
%   G = \bigl\{ \mathcal{G}^i_{\mathbf{q}^i} 
%   \mid i \in \mathcal{I}, \mathbf{q}^i \in \mathcal{D}^i \bigr\},
% $$
% as the collection of all solution maps for the entire range of PDE families and parameter values under consideration. An MOL approach seeks to approximate this collection $G$ with a single, unified model that is capable of providing accurate predictions for any $i \in \mathcal{I}$ and any $\mathbf{q}^i \in \mathcal{D}^i$. Consequently, MOL exhibits superior generalizability compared to an SOL approach, since it does not require retraining whenever $i$ or $\mathbf{q}^i$ changes.

% advection, wave, Klein-Gordon, Sine-Gordon k_tot=2, others k_tot=4. Klein-Gordon, Sine-Gordon num_choise_k=1, others num_choise_k=2

% \begin{itemize}
%     \item Forward 1-D time-dependent PDE
%     \item Periodic BC
%     \item IC: depends on the order of the derivative with respect to $t$ (e.g., $u_t$, $u_{tt}$). first derivative
%     \item 10 families of PDEs for training and testing under cases with normal/limited data, 3 families of PDEs for extrapolation or transfer learning. Generation of ICs. Numerical solvers.
%     \item SOL and MOL. Cited from LeMON: MOL can be categorized into two classes: (1) non-operator encoding and (2) operator encoding (incorporate an embedding of the operator (the equation, a label, etc.)). 
%     \item Multimodal Machine Learning (MML)

%     \item 
% \end{itemize}

\section{Methods}
\label{sec:methods}

In this section, we detail our modeling and training methodology. In Section~\ref{subsec:workflow_algorithm}, we introduce the PI-MFM framework and physics-informed training objective, defining the four loss terms (PDE, IC, IC$^\prime$, and data) and their weighted combination. In Section~\ref{subsec:discrete_pde_loss}, we then present the practical discrete formulation and a vectorized implementation for PDE-residual evaluation, where symbolic PDEs are parsed into expression trees and the required derivatives are computed to form the residual losses efficiently. Section~\ref{subsec:differentiation_methods} compares differentiation backends for physics losses, automatic differentiation (AD) (Section~\ref{subsec:ad}) versus finite difference method (FDM) (Section~\ref{subsec:fdm}), and explains how we leverage their complementary strengths in training.
Finally, Section~\ref{subsec:downstream_method} introduces physics-informed zero-shot fine-tuning to adapt the pretrained multi-operator model to unseen PDE families without labeled solutions.

\subsection{PI-MFM framework and training objective}
\label{subsec:workflow_algorithm}
% In this work, we aim to train a surrogate model $\mathcal{G}_{\theta}$ to approximate the multi-operator map $\mathcal{G}$. Specifically, our surrogate model maps an initial condition $ u_0 $ and a symbolic expression $ s(i,\mathbf{q}^i) $, which identifies a PDE with family index $ i $ and parameters $\mathbf{q}^i \in \mathcal{D}^i$, to the corresponding PDE solution $ u $.

\begin{figure}[htbp]
    \centering
    \includegraphics[width=\textwidth]{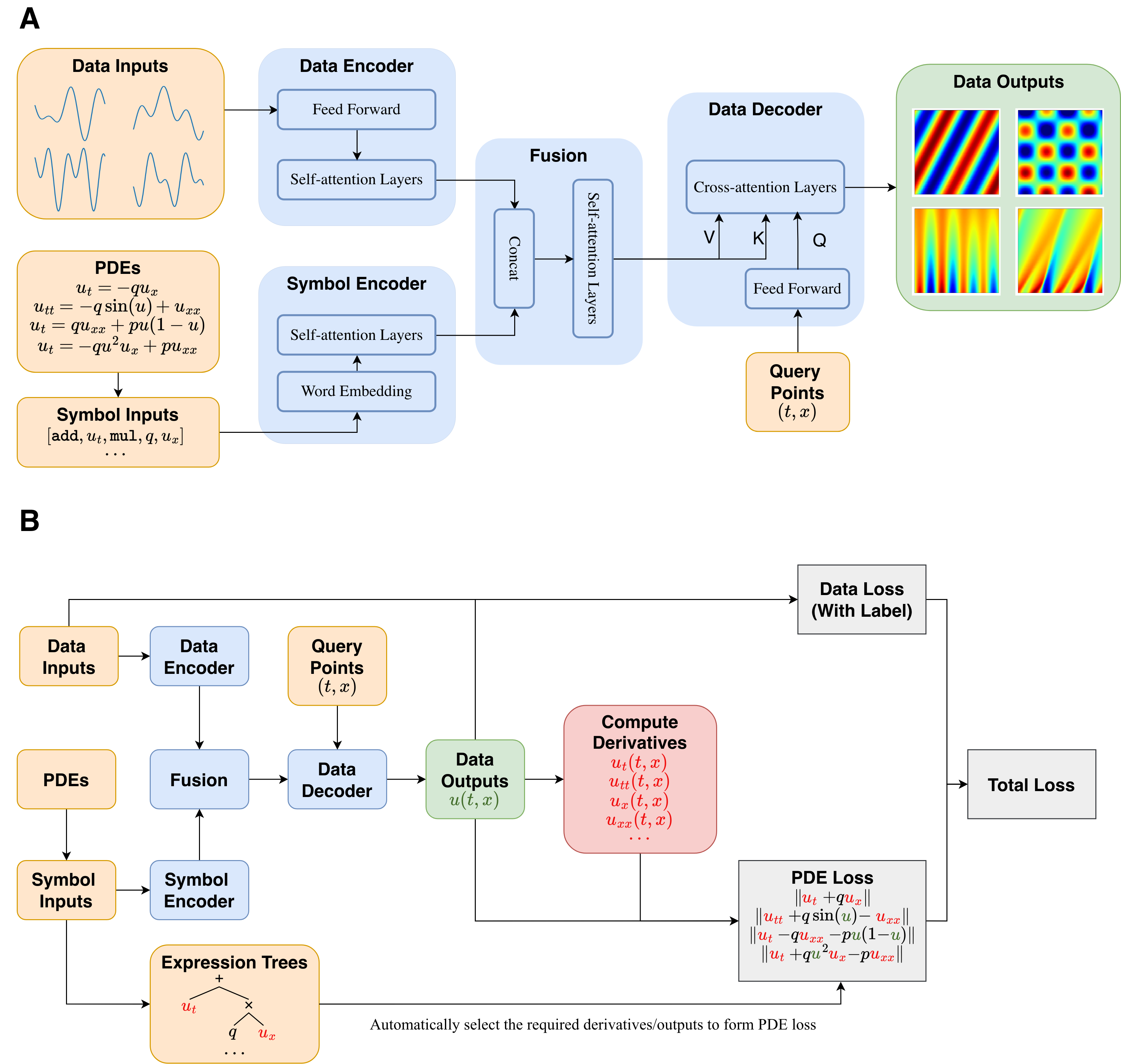}
    \caption{\textbf{Overview of the physics-informed multimodal foundation model (PI-MFM) framework for multi-operator learning (MOL).} (\textbf{A}) A representative example of PDE-encoding multi-operator learning framework based on PROSE~\cite{liu2024prose,sun2025towards}. It jointly encodes data inputs and symbolic PDE expressions, fuses these modalities, and decodes solution values at arbitrary query points $(t,x)$. (\textbf{B}) PI-MFM training workflow. Given a symbolic PDE, the framework automatically selects and computes the required derivatives of the PDE solution and assembles PDE-residual losses and data losses into a total loss. This workflow can be applied to any MFM that explicitly accepts PDE operator information. For simplicity, IC losses are omitted in the workflow and can be viewed as special cases of data or PDE losses. Periodic boundary conditions are enforced by augmenting the input coordinates with periodic features~\cite{lu2021physics,lu2022comprehensive}, so that periodicity is built into the network rather than imposed through an explicit BC loss term.}
    \label{fig:architecture}
\end{figure}

As discussed in Section~\ref{subsec:sol_mol}, our goal is to train a surrogate model, denoted by  $\mathcal{G}_\theta(u_0, s)$, that takes both the initial condition and the symbolic expression of the PDE as inputs and produces an approximation to the corresponding solution. 

\paragraph{Network architecture.}
Fig.~\ref{fig:architecture} summarizes the PI-MFM workflow. Specifically, Fig.~\ref{fig:architecture}A illustrates a PROSE-based implementation under our problem setup. The data stream (initial conditions) and the symbol stream (symbolic PDE expressions) are first embedded by their respective encoders, and the resulting embeddings are then processed by stacks of self-attention layers; a fusion module combines them into a context that supplies keys and values $(K,V)$ to a cross-attention data decoder, while the embedded query locations $(t,x)$ provide the queries $Q$. The decoder then predicts $u$ at arbitrary query points $(t,x)$. We use layer normalization for all self-attention and cross-attention layers, and the reason is explained in Section~\ref{subsec:ad}.

More generally, Fig.~\ref{fig:architecture}B abstracts this into an architecture-flexible design composed of a data encoder, a symbol encoder, a fusion module that combines the two streams, and a data decoder conditioned on the embedded query coordinates $(t,x)$. Each module may be implemented by any network that accepts the stated inputs and returns outputs with the required shapes. After embedding, the data and symbol inputs have shape $(B, N_{\mathrm{emb}})$, the query locations have shape $(M, N_{\mathrm{emb}})$, and the decoder outputs $u\in\mathbb{R}^{B\times M}$, where $N_{\mathrm{emb}}$ is the embedding width, $B$ denotes the number of PDEs in a batch, and $M$ denotes the number of query points per PDE.

\paragraph{Training objectives.}
The training losses in Fig.~\ref{fig:architecture}B are formed as follows. 
The PDE residual loss $\mathcal{L}_{\mathrm{PDE}}(\theta)$ uses the predicted solution together with its required spatiotemporal derivatives to evaluate the PDE residual:
$$\mathcal{L}_{\text{PDE}}(\theta) = %\frac{1}{|\mathcal{I}|}\sum_{i \in \mathcal{I}} 
\mathbb{E}_{s\sim\mathcal{S},\,u_0\sim\mathcal{U}_0}
\bigl\| \mathcal{F}(\mathcal{G}_{\theta}(u_0, s); s)\bigr\|^2.$$

The initial condition loss $\mathcal{L}_{\mathrm{IC}}(\theta)$ compares the decoder output at $t=0$ to the provided initial condition $u_0$:
$$\mathcal{L}_{\text{IC}}(\theta) = %\frac{1}{|\mathcal{I}|}\sum_{i \in \mathcal{I}} 
\mathbb{E}_{s\sim\mathcal{S},\,u_0\sim\mathcal{U}_0}
\bigl\|\mathcal{G}_{\theta}(u_0, s)(0,x)-u_0\bigr\|^2.$$
The second-order initial condition loss $\mathcal{L}_{\mathrm{IC}'}(\theta)$ imposes the additional initial condition $\frac{\partial u}{\partial t}(0, x) = 0$ for second-order-in-time PDEs:
$$\mathcal{L}_{\text{IC}^\prime}(\theta) = %\frac{1}{|\mathcal{I}^\prime|}\sum_{i \in \mathcal{I}^\prime} 
\mathbb{E}_{s\sim\mathcal{S}^\prime,\,u_0\sim\mathcal{U}_0}
\left\|\frac{\partial \mathcal{G}_{\theta}(u_0, s)}{\partial t}(0, x)\right\|^2.$$
where $\mathcal{S}' \subset \mathcal{S}$ denotes the subset of symbolic expressions corresponding to second-order-in-time PDEs.

The data prediction loss $\mathcal{L}_{\mathrm{data}}(\theta)$ requires the predicted solution to match the ground-truth solution:

$$\mathcal{L}_{\text{data}}(\theta) = %\frac{1}{|\mathcal{I}|}\sum_{i \in \mathcal{I}} 
\mathbb{E}_{s\sim\mathcal{S},\,u_0\sim\mathcal{U}_0}
\bigl\|\mathcal{G}_{\theta}(u_0, s)-\mathcal{G}(u_0, s)\bigr\|^2.$$
We combine these four loss terms to define the total training loss:
$$\mathcal{L}(\theta)= \omega_\text{PDE} \mathcal{L}_{\text{PDE}}(\theta) + \omega_\text{IC} \mathcal{L}_{\text{IC}}(\theta) + \omega_{\text{IC}^\prime}\mathcal{L}_{\text{IC}^\prime}(\theta) + \omega_\text{data} \mathcal{L}_{\text{data}}(\theta),$$
where $\omega_\text{PDE}$, $\omega_\text{IC}$, $\omega_{\text{IC}^\prime}$, and $\omega_\text{data}$ are the weights assigned to the PDE, IC, IC$^\prime$, and data loss terms, respectively.

\subsection{Discrete formulation and vectorized PDE loss computation}
\label{subsec:discrete_pde_loss}

\paragraph{Discrete formulation.}
In practical implementations, we work with finite training samples and discrete evaluation points. Accordingly, the loss functions are discretized as follows:
$$
\mathcal{L}_{\mathrm{PDE}}(\theta)
=
\frac{1}{|\mathcal{Z}^{\mathrm{PDE}}|}
\sum_{(s,u_0)\in\mathcal{Z}^{\mathrm{PDE}}}
\left[
\frac{1}{|\mathcal{T}_{s,u_0}^{\mathrm{PDE}}|}
\sum_{(t,x)\in\mathcal{T}_{s,u_0}^{\mathrm{PDE}}}
\bigl(
\mathcal{F}\bigl(
\mathcal{G}_{\theta}(u_0,s);\,s
\bigr)(t,x)
\bigr)^{2}
\right],
$$
$$
\mathcal{L}_{\mathrm{IC}}(\theta)
=
\frac{1}{|\mathcal{Z}^{\mathrm{IC}}|}
\sum_{(s,u_0)\in\mathcal{Z}^{\mathrm{IC}}}
\left[
\frac{1}{|\mathcal{T}_{s,u_0}^{\mathrm{IC}}|}
\sum_{x\in\mathcal{T}_{s,u_0}^{\mathrm{IC}}}
\bigl(
\mathcal{G}_{\theta}(u_0,s)(0,x)
-
u_0(x)
\bigr)^{2}
\right],
$$
$$
\mathcal{L}_{\mathrm{IC}^\prime}(\theta)
=
\frac{1}{|\mathcal{Z}^{\mathrm{IC^\prime}}|}
\sum_{(s,u_0)\in\mathcal{Z}^{\mathrm{IC^\prime}}}
\left[
\frac{1}{|\mathcal{T}_{s,u_0}^{\mathrm{IC}^\prime}|}
\sum_{x\in\mathcal{T}_{s,u_0}^{\mathrm{IC}^\prime}}
\biggl(
\bigl.
\frac{\partial \mathcal{G}_{\theta}(u_0,s)}{\partial t}
(0,x)
\biggr)^{2}
\right],
$$
$$
\mathcal{L}_{\mathrm{data}}(\theta)
=
\frac{1}{|\mathcal{Z}^{\mathrm{data}}|}
\sum_{(s,u_0)\in\mathcal{Z}^{\mathrm{data}}}
\left[
\frac{1}{|\mathcal{T}_{s,u_0}^{\mathrm{data}}|}
\sum_{(t,x)\in\mathcal{T}_{s,u_0}^{\mathrm{data}}}
\bigl(
\mathcal{G}_{\theta}(u_0,s)(t,x)
-
\mathcal{G}(u_0,s)(t,x)
\bigr)^{2}
\right],
$$
where $\mathcal{Z}^{\mathrm{PDE}}$, $\mathcal{Z}^{\mathrm{IC}}$, $\mathcal{Z}^{\mathrm{IC}^\prime}$, and $\mathcal{Z}^{\mathrm{data}}$ denote discrete training sets of $(s,u_0)$ samples used for PDE, IC, IC$^\prime$, and data loss terms, respectively. These sets may be chosen to be identical or distinct, depending on data availability and training strategy. For example, in Sections~\ref{subsec:sparse_labeled} and~\ref{subsec:partially_labeled}, we set $\mathcal{Z}^{\mathrm{PDE}} = \mathcal{Z}^{\mathrm{IC}} = \mathcal{Z}^{\mathrm{IC}^\prime} = \mathcal{Z}^{\mathrm{data}}$ and randomly resample these sets at every iteration. In contrast, in Section~\ref{subsec:small_func}, sparse function space data necessitate $\mathcal{Z}^{\mathrm{PDE}} = \mathcal{Z}^{\mathrm{IC}} = \mathcal{Z}^{\mathrm{IC}^\prime} \neq \mathcal{Z}^{\mathrm{data}}$.

The sets  $\mathcal{T}_{s,u_0}^{\mathrm{PDE}}$, $\mathcal{T}_{s,u_0}^{\mathrm{IC}}$, $\mathcal{T}_{s,u_0}^{\mathrm{IC^\prime}}$, and $\mathcal{T}_{s,u_0}^{\mathrm{data}}$ specify the discrete collections of spatiotemporal (for PDE and data terms) or spatial (for IC and IC$^{\prime}$ terms) points at which the corresponding loss terms are evaluated.  In particular, the choice of $\mathcal{T}_{s,u_0}^{\mathrm{data}}$ depends on data availability, while the selection of $\mathcal{T}_{s,u_0}^{\mathrm{PDE}}$ is flexible. Any set of points in the spatiotemporal domain may be chosen for evaluating the PDE residual. Notably, the choice of $\mathcal{T}_{s,u_0}^{\mathrm{PDE}}$ can significantly affect training performance, as discussed in Section~\ref{subsec:resample_num_points}. In our implementation, we construct $\mathcal{T}_{s,u_0}^{\mathrm{PDE}}$ by randomly sampling collocation points within the spatiotemporal domain of the PDE, and we resample these points at every training iteration.

\begin{algorithm}[htbp]
\caption{\textbf{Vectorized PDE loss computation.}}
\label{alg:pde_loss_batch}
\begin{algorithmic}[1]
\Require network parameters $\theta$; minibatch $\{(u_{0,b}, s_b)\}_{b=1}^{B} \subset \mathcal{Z}^{\mathrm{PDE}}$; shared collocation point set $TX = \{(t_m,x_m)\}_{m=1}^{M}$ (sampled randomly in domain and resampled every iteration); derivative backend (AD or FDM)
\Ensure PDE loss $\mathcal{L}_{\mathrm{PDE}}(\theta)$ for the minibatch

\AlgBlockComment{\# Compute network outputs on collocation points}
\State Let $\mathbf{U}_0 := (u_{0,1},\ldots,u_{0,B})$ and $\mathbf{S} := (s_1,\ldots,s_B)$, where each $s_b$ is a symbolic input encoded as a token sequence in Polish notation
\State Perform a single batched forward pass of the model on all collocation points and set $U \gets \mathcal{G}_\theta(\mathbf{U}_0,\mathbf{S})(TX) \in \mathbb{R}^{B\times M}$, so that $U_{b,m} = \hat u_\theta(u_{0,b}, s_b)(t_m, x_m)$ for item $b$ and collocation point $m$

\AlgBlockComment{\# Compute required derivatives using AD or FDM}
\State From the batch of token sequences $\mathbf{S}$, form the union of all derivative types
       required in the current minibatch,
       $\mathcal{D} \subseteq \{u_t, u_{tt}, u_x, u_{xx}, \ldots\}$
\State For each derivative $d\in\mathcal{D}$, let $U_d\in\mathbb{R}^{B\times M}$ store its values at all items and collocation points (e.g., $(U_{u_t})_{b,m}=\partial u_{b,m}/\partial t_m$ and $(U_{u_{xx}})_{b,m}=\partial^2 u_{b,m}/\partial x_m^2$).

\If{using AD}
    \State Use vectorized calls to the AD backend to compute $\{U_d\}_{d\in\mathcal{D}}$ at all items and collocation points

\ElsIf{using FDM}
   \State Construct the stacked shifted collocation blocks needed for every $d\in\mathcal{D}$, e.g.,
          $TX \pm \Delta t\,\mathbf e_t$, $TX \pm \Delta x\,\mathbf e_x$, etc.
   \State Perform one vectorized forward pass of $\mathcal{G}_\theta$ on the stacked blocks and
          assemble finite differences to obtain $\{U_d\}_{d\in\mathcal{D}}$
\EndIf

\AlgBlockComment{\# Compute PDE residual loss}
\State Pack $\mathbf{Z} \in \mathbb{R}^{B \times M \times C}$ with $C = 1 + |\mathcal{D}|$ channels, placing $U$ in the first channel and the tensors $\{U_d\}_{d\in\mathcal{D}}$ in a fixed order in the remaining channels
\For{$b = 1$ to $B$}
   \State Let $\mathbf{Z}_b \in \mathbb{R}^{M \times C}$ be the slice of $\mathbf{Z}$ corresponding to item $b$
   \State Parse the token sequence $s_b$ into an expression tree $t_b$
   \State Recursively evaluate the expression tree $t_b$ on $\mathbf{Z}_b$
          to obtain the residual vector $r_b \in \mathbb{R}^{M}$ with entries
          $r_{b,m} = \mathcal{F}(\mathcal{G}_\theta(u_{0,b},s_b); s_b)(t_m,x_m), \quad m=1,\ldots,M$
\EndFor
\State Compute the batch PDE loss
       $\mathcal{L}_{\mathrm{PDE}}(\theta)
       \gets \frac{1}{BM}\sum_{b=1}^{B}\sum_{m=1}^{M} r_{b,m}^{2}$
\State \Return $\mathcal{L}_{\mathrm{PDE}}(\theta)$
\end{algorithmic}
\end{algorithm}

% \paragraph{Algorithm for PDE loss computation.}
% Let $B=|\mathcal{Z}^{\mathrm{PDE}}|$ denote the number of PDE items in the batch and $M=|\mathcal{T}^{\mathrm{PDE}}_{s,u_0}|$ the number of collocation points per item (the same $M$ for all items). For each item $b$, the Polish-notation token sequence is parsed into an expression tree $s_b$. We assemble a batch tensor $\mathbf{Z}\in\mathbb{R}^{B\times M\times C}$ whose $b$-th slice $\mathbf{Z}_b\in\mathbb{R}^{M\times C}$ holds the $C$ channels $[u, u_t, u_{tt}, u_x, u_{xx}, \ldots]$ evaluated at the $M$ points for item $b$. Leaves of $s_b$ select columns of $\mathbf{Z}_b$ or return broadcast constants; internal nodes apply element-wise arithmetic (such as addition and multiplication) and unary operations (such as exponential and sine). Since each leaf returns an $M$-vector, evaluating $s_b$ yields a residual vector $r_b\in\mathbb{R}^{M}$. The PDE loss is the average of the residuals over both axes: the $B$ items and the $M$ collocation points. This loss formulation does not depend on how $\mathbf{Z}$ is produced, whether by automatic differentiation (AD, see Section~\ref{subsec:ad}), finite difference methods (FDM, see Section~\ref{subsec:fdm}), or other approaches, and it does not rely on the structure of the neural network: any MOL foundation model that accepts operator information (e.g., symbolic tokens) can be trained or adapted with the same physics losses without architectural changes. The algorithm for batch PDE loss computation is shown in Algorithm~\ref{alg:pde_loss_batch}.

\paragraph{Algorithm for PDE loss computation.}
Let $B=|\mathcal{Z}^{\mathrm{PDE}}|$ denote the number of PDE items in the minibatch and let $\mathcal{T}^{\mathrm{PDE}}_{s_b,u_{0,b}}=\{(t_m,x_m)\}_{m=1}^{M}$ denote the collocation set at which we enforce the PDE residual for item $(s_b,u_{0,b})$. In the batched implementation of Algorithm~\ref{alg:pde_loss_batch}, we use the same number $M$ of collocation points for every item and typically share a common collocation set $TX$ across items, i.e., $\mathcal{T}^{\mathrm{PDE}}_{s_b,u_{0,b}} = TX$, which is sampled randomly in domain and resampled every iteration. For each item $b$, we represent the PDE operator by a symbolic input $s_b$ encoded as a token sequence in Polish notation, and we parse $s_b$ once into an expression tree $t_b$ used for PDE loss evaluation.

We first evaluate the model in a single batched forward pass on the shared collocation set $TX=\{(t_m,x_m)\}_{m=1}^M$ to obtain $U$ for all minibatch items $b$ and collocation points $m$. From the minibatch of token sequences $\mathbf{S}$, we then form the union of all derivative types required by the PDE residuals in the current minibatch, $\mathcal{D}\subseteq\{u_t,u_{tt},u_x,u_{xx},\ldots\}$. For each $d\in\mathcal{D}$, we denote by $U_d\in\mathbb{R}^{B\times M}$ the corresponding derivative values evaluated at all items and collocation points (e.g., $(U_{u_t})_{b,m}=\partial u_{b,m}/\partial t_m$ and $(U_{u_{xx}})_{b,m}=\partial^2 u_{b,m}/\partial x_m^2$). When using AD, we compute $\{U_d\}_{d\in\mathcal{D}}$ via vectorized calls to the AD backend at all items and collocation points (Section~\ref{subsec:ad}). In contrast, FDM runs a single forward pass on one enlarged query batch obtained by concatenating several shifted copies of the original collocation set (e.g., $TX \pm \Delta t\,\mathbf e_t$, $TX \pm \Delta t\,\mathbf e_x$, $TX$), and then forms the desired finite differences from these outputs (Section~\ref{subsec:fdm}). This produces a tensor $\mathbf{Z}\in\mathbb{R}^{B\times M\times C}$ whose $b$-th slice $\mathbf{Z}_b\in\mathbb{R}^{M\times C}$ holds the $C=1+|\mathcal{D}|$ channels $[u, u_t, u_{tt}, u_x, u_{xx},\ldots]$ evaluated at the $M$ points for the $b$-th PDE.

Leaves of the expression tree $t_b$ (operands) select channels from $\mathbf{Z}_b$ or return constants, while internal nodes (operators) apply element-wise
mathematical operations. Recursively evaluating $t_b$ on $\mathbf{Z}_b$ therefore produces the residual vector $r_b\in\mathbb{R}^M$ at the $M$ shared
collocation points. The minibatch PDE loss is then computed as the mean squared residual over both the instance index $b$
and point index $m$. Crucially, once $\mathbf{Z}$ is formed, evaluating $t_b$ involves only cheap tensor arithmetic
and never calls the neural network or the derivative backend again. As a result, the runtime of PDE-loss evaluation is dominated by the single
derivative-generation stage per minibatch, and the same procedure applies to any multimodal foundation model that accepts operator information
(e.g., symbolic tokens) without architectural changes.

\subsection{Analysis and choice of differentiation methods}
\label{subsec:differentiation_methods}

Physics-informed operator learning typically solves PDEs by incorporating physical constraints, such as the governing equations, initial conditions, and boundary conditions, into the loss function. A crucial step in this process is computing the derivatives of the neural network output with respect to its input variables (such as spatial coordinates and time), which are required to evaluate the PDE residual and any initial or boundary conditions involving derivatives. The most commonly used methods for this purpose are the automatic differentiation (AD) and finite difference method (FDM)~\cite{raissi2019physics,baydin2018automatic,pang2019fpinns}.

\subsubsection{Optimal automatic differentiation (AD)}
\label{subsec:ad}

% \textcolor{blue}{
% We compare forward- and reverse-mode AD for evaluating physics residuals in multi-operator learning. Let the network output for one example be $u=\mathcal{G}_\theta(u_0,s)(t,x)$. For a batch of $B$ PDE instances $\{s_b\}_{b=1}^B$ and $M$ coordinates per instance, collect inputs as $Z=\{z_{b,m}\}_{b=1..B,\;m=1..M}$ with $z_{b,m}=(t_{bm},x_{bm},s_b,u_{0,b})$, and define the batched mapping $F(Z)=\{u_{b,m}\}_{b,m}$, where $u_{b,m}=G_\theta(u_{0,b},s_b)(t_{bm},x_{bm})$. Because standard network layers do not mix across the batch dimension, the Jacobian of $F$ with respect to the per-point coordinates $(t,x)$ is block diagonal across $(b,m)$. Writing $J=\partial F/\partial(t,x)\in\mathbb{R}^{(BM)\times(2BM)}$ and $J_{b,m}=\begin{bmatrix}\partial u_{b,m}/\partial t&\partial u_{b,m}/\partial x\end{bmatrix}\in\mathbb{R}^{1\times 2}$, we have $J=\mathrm{diag}(\{J_{b,m}\}_{b,m})$ with no off-diagonal couplings between distinct $(b,m)$. This structure is what allows forward mode to commute with batching.}

We compare forward- and reverse-mode AD for evaluating physics residuals in multi-operator learning. For a single PDE instance, the model predicts the solution value at a coordinate $(t,x)$ as
\[
u=\mathcal{G}_\theta(u_0,s)(t,x),
\]
where $(u_0,s)$ denote the initial condition and a symbolic PDE descriptor. For a minibatch of $B$ PDE instances
\[
Z:=\{(u_{0,b},s_b)\}_{b=1}^B,
\]
we evaluate the model on a shared set of $M$ query coordinates
\[
T:=\{(t_m,x_m)\}_{m=1}^M,
\qquad
t=(t_1,\ldots,t_M)^\top\in\mathbb{R}^M,\ \ x=(x_1,\ldots,x_M)^\top\in\mathbb{R}^M,
\]
and collect the outputs
\[
u_{b,m}=\mathcal{G}_\theta(u_{0,b},s_b)(t_m,x_m).
\]
Stacking all outputs defines a matrix-valued mapping
\[
F(T;Z)\in\mathbb{R}^{B\times M},
\qquad
[F(T;Z)]_{b,m}=u_{b,m}.
\]

When writing Jacobians, we flatten $F(T;Z)$ as $\operatorname{vec}(F(T;Z))\in\mathbb{R}^{BM}$. We focus on partial derivatives with respect to the
coordinate variables $(t,x)$, treating $Z$ as fixed. The resulting Jacobian is
\[
J=\frac{\partial\,\operatorname{vec}(F(T;Z))}{\partial\,[t;x]}
\in\mathbb{R}^{(BM)\times(2M)}.
\]
If the forward pass is pointwise in the query index $m$ (i.e., there is no mixing across different $m$), then changing $(t_m,x_m)$ affects only the
outputs at the same index $m$, namely $\{u_{b,m}\}_{b=1}^B$. Consequently, $J$ has nonzero entries only within each point index $m$: the $B$ outputs $\{u_{b,m}\}_{b=1}^B$ depend only on the two variables
$(t_m,x_m)$ and are independent of $(t_{m'},x_{m'})$ for $m'\neq m$. Equivalently, $J$ is block diagonal across $m$, with one $B\times 2$ block per
query point. The row corresponding to output $(b,m)$ is
\[
J_{b,m}=\begin{bmatrix}\partial u_{b,m}/\partial t_m & \partial u_{b,m}/\partial x_m\end{bmatrix}\in\mathbb{R}^{1\times 2}.
\]
the Jacobian has no couplings between indices $m'\neq m$. This block-diagonal structure is the key reason forward-mode AD can be applied in a fully
vectorized manner.

% \textcolor{blue}
% {Forward mode computes Jacobian-vector products (JVP). Let $V\in\mathbb{R}^{B\times M\times 2}$ collect the tangent directions aligned with the $(t,x)$ inputs, with $V_{b,m,:}=v_{b,m}$. A single vectorized forward sweep returns a tensor in $\mathbb{R}^{B\times M}$, whose $(b,m)$ entry is
% $$
% \big[\mathrm{JVP}(F;Z,V)\big]_{b,m}=J_{b,m}\,v_{b,m}
% =\frac{\partial u_{b,m}}{\partial t}\,V_{b,m,1}\;+\;\frac{\partial u_{b,m}}{\partial x}\,V_{b,m,2}.
% $$
% Choosing $v_{b,m}=e_t=(1,0)$ yields all $\{\partial u_{b,m}/\partial t\}_{b,m}$ in one pass; choosing $v_{b,m}=e_x=(0,1)$ yields all $\{\partial u_{b,m}/\partial x\}_{b,m}$. Higher-order and mixed derivatives follow by nesting JVPs: compute $u_x$ as a primal via one JVP, then apply a second JVP with $e_x$ or $e_t$ to obtain $u_{xx}$ or $u_{tx}$. The key complexity property is that cost scales with the number of input directions per point (two for $(t,x)$) and the derivative order, but is essentially independent of the number of points $BM$. This makes forward mode natural for physics-informed objectives that require derivatives at many coordinates simultaneously.}

Forward mode evaluates Jacobian--vector products (JVPs) with respect to the shared coordinate variables. Since all PDE instances share the same $M$
query coordinates, we specify one tangent direction per coordinate pair $(t_m,x_m)$. Let
\[
V\in\mathbb{R}^{M\times 2},\qquad V_{m,:}=v_m,
\]
collect these tangents. A single vectorized forward-mode sweep returns a tensor in $\mathbb{R}^{B\times M}$ whose $(b,m)$ entry is
\[
\big[\mathrm{JVP}(F; (t,x), V)\big]_{b,m}
= J_{b,m}\,v_m
=\frac{\partial u_{b,m}}{\partial t_m}\,V_{m,1}\;+\;\frac{\partial u_{b,m}}{\partial x_m}\,V_{m,2}.
\]
Choosing $v_m=e_t=(1,0)$ for all $m$ yields all $\{\partial u_{b,m}/\partial t_m\}_{b,m}$ in one pass; choosing $v_m=e_x=(0,1)$ yields all
$\{\partial u_{b,m}/\partial x_m\}_{b,m}$ in one pass.

Higher-order and mixed derivatives follow by nesting JVPs. For example, first obtain $u_x=\{\partial u_{b,m}/\partial x_m\}_{b,m}$ via one JVP with
$e_x$, then apply a second JVP to the mapping $(t,x)\mapsto u_x$ with $e_x$ or $e_t$ to obtain $u_{xx}$ or $u_{tx}$, respectively. The key
complexity property is that the number of AD sweeps scales with the number of coordinate directions (two for $(t,x)$) and the derivative order, while
each sweep is applied to all $B\times M$ outputs in parallel.

% \textcolor{blue}{Reverse mode computes vector–Jacobian products (VJP). Given cotangents $W\in\mathbb{R}^{B\times M}$ aligned with $F(Z)$, a single reverse sweep returns a tensor in $\mathbb{R}^{B\times M \times 2}$ whose $(b,m)$ entry is
% $$
% \big[\mathrm{VJP}(F;Z,W)\big]_{b,m,:}
% =J_{b,m}^{\top}W_{b,m}
% =\begin{bmatrix}
% \frac{\partial u_{b,m}}{\partial t}\,W_{b,m}\\[2pt]
% \frac{\partial u_{b,m}}{\partial x}\,W_{b,m}
% \end{bmatrix}.
% $$
% (Flattening the batch axes gives the equivalent form $J^{\top}\operatorname{vec}(W)\in\mathbb{R}^{2BM}$.) Let $E_{b,m}\in\mathbb{R}^{B\times M}$ be the one-hot tensor with $(E_{b,m})_{i,j}=\delta_{i,b}\delta_{j,m}$. Choosing $W=E_{b,m}$ isolates the single row $J_{b,m}^{\top}$. To obtain all $BM$ rows one must evaluate reverse mode for $BM$ cotangents, either by looping or by vectorizing the reverse-mode Jacobian evaluation; in both cases, the work scales linearly with the number of outputs, and the backward sweep must retain activations. For higher-order derivatives, differentiating through a backward pass (reverse-over-reverse or forward-over-reverse) further increases computation and memory.}

Reverse mode evaluates vector--Jacobian products (VJPs) with respect to the shared coordinate variables. Given cotangents
$W\in\mathbb{R}^{B\times M}$ aligned with $F(t,x;Z)$, a single reverse sweep returns gradients with respect to $(t,x)$:
\[
\mathrm{VJP}(F;(t,x),W)
=
J^{\top}\operatorname{vec}(W)
\in\mathbb{R}^{2M},
\qquad
J\in\mathbb{R}^{(BM)\times(2M)}.
\]
Equivalently, writing the result as $(g_t,g_x)\in\mathbb{R}^{M}\times\mathbb{R}^{M}$, the $m$th entries are
\[
(g_t)_m=\sum_{b=1}^B \frac{\partial u_{b,m}}{\partial t_m}\,W_{b,m},
\qquad
(g_x)_m=\sum_{b=1}^B \frac{\partial u_{b,m}}{\partial x_m}\,W_{b,m}.
\]
Thus, a single VJP aggregates contributions across PDE instances $b$ at each shared coordinate index $m$.

To isolate the coordinate-gradient contribution of a single output entry $(b,m)$, one may choose the one-hot cotangent
$W=E_{b,m}\in\mathbb{R}^{B\times M}$ with $(E_{b,m})_{i,j}=\delta_{i,b}\delta_{j,m}$, which yields
\[
(g_t)_m=\frac{\partial u_{b,m}}{\partial t_m},\qquad (g_x)_m=\frac{\partial u_{b,m}}{\partial x_m},
\]
and zero for all $m'\neq m$. However, because the coordinates $(t_m,x_m)$ are shared across $b$, a single reverse sweep computes only the aggregated coordinate gradients $(g_t,g_x)=J^\top\operatorname{vec}(W)$ for a chosen cotangent $W$, rather than the full collection of entrywise derivatives $\{(\partial u_{b,m}/\partial t_m,\partial u_{b,m}/\partial x_m)\}_{b,m}$. Extracting all $BM$ per-output coordinate derivatives via reverse mode would require $BM$ such one-hot cotangents (by looping or vectorizing Jacobian extraction implemented via PyTorch function \texttt{torch.vmap}), requiring $BM$ VJPs. Each sweep must backpropagate through the forward computation, which typically requires storing or recomputing intermediate activations; repeating this $BM$ times leads to substantial time and memory overhead in practice. For higher-order derivatives, differentiating through a backward pass further increases computation and memory.

% \textcolor{blue}{In practice, we evaluate per-coordinate derivatives with forward mode and compute parameter gradients with a single reverse sweep after assembling a scalar objective. Forward mode produces the coordinate derivatives required by the physics losses over all $B\times M$ points in one JVP pass, with higher-order and mixed derivatives obtained by nesting JVPs; this one-shot behavior holds because batch items do not interact and the Jacobian with respect to $(t,x)$ is block diagonal. Any operation that mixes examples across the batch dimension, such as training-mode batch normalization, cross-sample attention, or batch-wise aggregation, introduces off-diagonal couplings and breaks this property. Once a scalar loss $\mathcal{L}$ is formed, a single reverse-mode sweep computes $\nabla_\theta \mathcal{L}$, preserving reverse mode’s efficiency for parameter gradients.}

In practice, we compute coordinate derivatives for the physics residuals using a constant number of forward-mode sweeps (two directions for first-order $(t,x)$ derivatives), with higher-order and mixed derivatives obtained by nesting JVPs. This batched behavior relies on the forward pass being pointwise in the query index $m$, so that the Jacobian with respect to $(t,x)$ has no couplings between indices $m'\neq m$. Any operation that mixes different query indices introduces off-diagonal couplings and breaks this property. Examples include training-mode batch normalization that aggregates statistics across multiple query indices $m$, attention across query points, and other aggregations that couple several values of $m$. In our implementation, we use layer normalization (rather than batch normalization), which normalizes each $(b,m)$ independently across feature dimensions and therefore avoids introducing such cross-$m$ couplings. Once a scalar loss $\mathcal{L}$ is formed from these residual terms, a single reverse-mode sweep through the resulting computation graph computes $\nabla_\theta\mathcal{L}$, preserving reverse mode's efficiency for parameter gradients.

\subsubsection{Finite difference method (FDM): Optimal step size and floating-point format}
\label{subsec:fdm}

Finite difference methods (FDM) approximate derivatives using function values at neighboring points. These methods require selecting a small increment for each coordinate and are typically derived using Taylor expansions. Finite difference schemes are conceptually straightforward and do not require explicit knowledge of the analytic form of the function. In this work, we use central difference formulas and evaluate them using various floating-point formats, including float64, float32, float16, and bfloat16.

The choice of spatial and temporal discretization step sizes plays a critical role in determining both numerical accuracy and stability. Importantly, in PI-MFM the finite differences are applied using small pointwise neighborhoods of query points around each collocation point where the physics loss is evaluated, rather than on a fixed solver grid. Because the surrogate can be queried at arbitrary coordinates $(t,x)$, the neighboring query points used by finite differences (e.g., $(t,x\pm\Delta x)$ and $(t\pm\Delta t,x)$) can be chosen on any mesh: uniform or nonuniform, and even on collocation sets that are resampled or adaptively refined during training. In other words, the “grid” is not a solver discretization but the collocation set used to evaluate the physics loss, so $\Delta x,\Delta t$ need not correspond to a fixed global lattice.

To quantitatively analyze finite difference errors, consider the central difference approximation for the second derivative. We focus on the second derivative of $u$ with respect to $x$:
$$
u''(x)\approx \frac{u(x+\Delta x)-2u(x)+u(x-\Delta x)}{\Delta x^2}.
$$
which has a truncation error of order $\mathcal{O}(\Delta x^2)$.
The total numerical error, $E$, consists of the truncation error ($E_{\text{trunc}}$) from the finite difference approximation and the round-off error $(E_{\text{round}})$ due to finite-precision arithmetic.

% reomove deduction
\paragraph{Truncation error.}  
By expanding $u(x+\Delta x)$ and $u(x-\Delta x)$ via Taylor series around $x$, we obtain
\begin{align*}
u(x+\Delta x)&=u(x)+u'(x)\Delta x+\frac{u''(x)}{2}\Delta x^2+\frac{u'''(x)}{6}\Delta x^3+\frac{u^{(4)}(x)}{24}\Delta x^4+O(\Delta x^5),\\
u(x-\Delta x)&=u(x)-u'(x)\Delta x+\frac{u''(x)}{2}\Delta x^2-\frac{u'''(x)}{6}\Delta x^3+\frac{u^{(4)}(x)}{24}\Delta x^4+O(\Delta x^5).\end{align*}
Thus, the leading-order truncation error is
$$ E_{\text{trunc}} = \frac{u(x+\Delta x)-2u(x)+u(x-\Delta x)}{\Delta x^2} -u''(x) \approx \frac{u^{(4)}(x)}{12}\Delta x^2\propto u^{(4)}(x)\Delta x^2. $$

\paragraph{Round-off error.} 
For floating-point arithmetic, each time we evaluate $u(x)$, the result has a small relative error, denoted by $\epsilon$, which depends on the floating-point precision used (e.g., $\epsilon =2^{-52} \approx 2.22 \times 10^{-16}$ for float64, $\epsilon =2^{-23} \approx 1.19 \times 10^{-7}$ for float32, $\epsilon =2^{-10} \approx 9.77 \times 10^{-4}$ for float16, and $\epsilon =2^{-7} \approx 7.81 \times 10^{-3}$ for bfloat16). The absolute error in each function value is therefore about $u(x)\epsilon$. Since $u(x+\Delta x)$ and $u(x-\Delta x)$ are both close to $u(x)$, after dividing by $\Delta x^2$, the round-off error is estimated as
$$E_{\text{round}}\propto \frac{u(x)\epsilon}{\Delta x^2}.$$

Thus, the total error in the finite difference approximation can be expressed as
$$ E_2 \approx A_2u^{(4)}(x)\Delta x^2 + B_2\frac{u(x)\epsilon}{\Delta x^2},$$
where $A_2$ and $B_2$ are constants. 
Similarly, for $n$-th order central difference, the combined truncation and round-off error is
$$ E_n \approx A_nu^{(n+2)}(x)\Delta x^2 + B_n\frac{u(x)\epsilon}{\Delta x^n},$$
where $A_n$ and $B_n$ are constants depending on $n$.

% Similarly, for first-order derivatives $u'(x)\approx \frac{u(x+\Delta x)-u(x-\Delta x)}{2\Delta x}$:
% $$ E_{\text{total}} \approx \frac{u'''(x)}{6}\Delta x^2 + \frac{u(x)\epsilon}{2\Delta x}, $$

% for third-order derivatives $u'''(x) \approx \frac{u(x+2\Delta x)-2u(x+\Delta x)+2u(x-\Delta x)-u(x-2\Delta x)}{2\Delta x^3}$:
% $$
% E_{total} \approx \frac{u^{(5)}(x)}{4}\Delta x^2 + \frac{u(x)\epsilon}{2\Delta x^3},
% $$

% and for fourth-order derivatives $u^{(4)}(x) \approx \frac{u(x+2\Delta x)-4u(x+\Delta x)+6u(x)-4u(x-\Delta x)+u(x-2\Delta x)}{\Delta x^4}$:
% $$
% E_{total} \approx \frac{u^{(6)}(x)}{6}\Delta x^2 + \frac{u(x)\epsilon}{\Delta x^4}
% $$

The truncation error is proportional to $\Delta x^2$, meaning that as $\Delta x$ decreases, the truncation error decreases rapidly. For the $n$-th order derivative, the truncation error is also proportional to $u^{(n+2)}(x)$, indicating that functions with rapid oscillations or steep gradients tend to exhibit higher truncation error.
In contrast, the round-off error is proportional to $\Delta x^{-n}$, so as $\Delta x$ becomes smaller, the round-off error becomes larger. In particular, when $\Delta x < 1$, the round-off error increases dramatically as the order $n$ increases.

For given values of $\epsilon$, $n$, $u(x)$, and $u^{(n+2)}(x)$, the optimal choice of $\Delta x$ that minimizes the total error is
\begin{equation}
\label{eq:opt_dx}
\Delta x_{\text{optimal}} \propto \left(\frac{u(x)\epsilon}{u^{(n+2)}(x)}\right)^{\frac{1}{n+2}}=\left(\frac{\epsilon}{R_n(x)}\right)^{\frac{1}{n+2}},
\end{equation}
where $R_n(x)=\frac{u^{(n+2)}(x)}{u(x)}$. The corresponding minimum total error is
$$E_{\text{min}}\propto (u(x)\epsilon)^{\frac{2}{n+2}}(u^{(n+2)}(x))^{\frac{n}{n+2}},
$$
and the relative error can be expressed as 

\begin{equation}
\label{eq:min_error}
\frac{E_{\text{min}}}{u(x)}\propto \epsilon^{\frac{2}{n+2}}R_n(x)^{\frac{n}{n+2}}.
\end{equation}

% emphasize why it is important to analyze this for MOL (figure 7 U shape)
% add the description when n=2
Eqs.~\eqref{eq:opt_dx} and~\eqref{eq:min_error} reveal important considerations for applying FDM within physics-informed multi-operator learning. Both the optimal step size $\Delta x_{\text{optimal}}$ and the minimum total error $E_{\text{min}}$ depend on $\epsilon$, $n$, and $R_n(x)$. 
In the SOL setting, $n$ is fixed and the solutions typically share similar characteristics, so $R_n(x)$ is expected to lie in a comparable range.
In contrast, for MOL, $n$ varies across different PDE families, and the range of $R_n(x)$ can differ significantly across these families. For example, conservation law families often exhibit much larger $R_n(x)$ than other families.

Unlike classical FDM time-stepping solvers that advance $u$ by discretizing the PDE, here finite differences are used only to evaluate derivatives inside the physics loss. Consequently, the solution error is not linearly tied to the derivative error; small derivative inaccuracies typically act like mild noise in the residual and might not degrade predictive accuracy. Moreover, Eqs.~\eqref{eq:opt_dx} and~\eqref{eq:min_error} imply that the optimal step size depends on the derivative order $n$ and the local ratio $R_n(x)=u^{(n+2)}(x)/u(x)$, so different derivatives (e.g., $u_t,u_x,u_{xx}$) generally prefer different $\Delta x$ values, and no single step size is uniformly optimal across PDE families. 

Because $\epsilon$ enters Eqs.~\eqref{eq:opt_dx} and \eqref{eq:min_error} alongside $n$ and $R_n(x)$, the trade-off between truncation and round-off errors is problem-dependent, and float64, float32, float16, or bfloat16 may each be preferable in different regimes. We therefore select step sizes and precisions empirically via numerical experiments (Section~\ref{subsec:accuracy_efficiency}).

% Unlike FDM numerical solver which directly uses finite difference, physics-informed methods just use finite difference in PDE loss. The error of the prediction is not proportional to the error of the derivatives. The small error in derivatives and PDE loss can be regarded as small noise and might not affect accuracy. We cannot determine the overall best step size since each derivative has its own best step size.
% We cannot directly determine when it is proper to use float64, float32, float16, or bfloat16.
% Thus, numerical experiments are necessary (See Section 4.5).

% first- and second-derivatives are most common in PDEs. Thus, we mainly explore those kind of PDEs. We also explore one 3rd-order and one 4th-order PDE.

\subsection{Downstream task: Solving unseen PDE families by zero-shot physics-informed fine-tuning}
\label{subsec:downstream_method}

We introduce zero-shot fine-tuning for PDE operator learning: starting from a physics-informed, multi-operator foundation model $\mathcal{G}_{\theta}$ pretrained on a diverse set of PDE families, we adapt it to a new PDE family (a different PDE class rather than a coefficient change of a seen family) without any labeled input-output pairs. Adaptation is driven solely by the governing physics (the PDE residual and IC/BC information), so the model learns the new family from its equation rather than from example solutions. We treat this adaptation as a downstream task analogous to fine-tuning in language/vision foundation models, but performed with physics-only supervision. To our knowledge, no prior multi-operator foundation model conducts physics-only zero-shot fine-tuning to an entire unseen PDE family. Existing works either do prediction without fine-tuning~\cite{yang2023context,sun2025towards,sun2024lemon}, rely on few-shot supervised fine-tuning~\cite{sun2024lemon,zhu2023reliable}, or use a hand-crafted PDE loss for each target PDE (the PDE is not provided as model input)~\cite{zhang2024deeponet}.

We fine-tune on an unseen PDE family with index $i_\star$. For any instance in this family we write the symbolic expression as $s(i_\star,q)$ with $q\in D^{i_\star}$; after first mention we denote it simply by $s$. Let
$$
S^{i_\star}=\{\,s(i_\star,q): q\in D^{i_\star}\,\}.
$$
With the initial condition $u_0\sim\mathcal{U}_0$, the physics-only losses are
$$
\mathcal{L}_{\mathrm{PDE}}^{(i_\star)}(\theta)
=\mathbb{E}_{s\sim S^{i_\star},\,u_0\sim\mathcal{U}_0}\;
\big\|\,\mathcal{F}\!\big(\mathcal{G}_\theta(u_0,s);\,s\big)\big\|^2,
$$
$$
\mathcal{L}_{\mathrm{IC}}^{(i_\star)}(\theta)
=\mathbb{E}_{s\sim S^{i_\star},\,u_0\sim\mathcal{U}_0}\;
\big\|\,\mathcal{G}_\theta(u_0,s)(0,\cdot)-u_0\big\|^2.
$$
Since all members of a family share the same time order, we define the family-level indicator
$\mathbf{1}^{\,i_\star}_{\mathrm{2nd}}\in\{0,1\}$ (1 if the family is second-order in time). When $\mathbf{1}^{\,i_\star}_{\mathrm{2nd}}=1$, we also include
$$
\mathcal{L}_{\mathrm{IC}'}^{(i_\star)}(\theta)
=\mathbb{E}_{s\sim S^{i_\star},\,u_0\sim\mathcal{U}_0}\;
\big\|\,\partial_t \mathcal{G}_\theta(u_0,s)(0,\cdot)\big\|^2.
$$
Let $\theta_{\text{pre}}$ denote the pretrained model parameters obtained in Section~\ref{subsec:workflow_algorithm}. We initialize $\theta\leftarrow\theta_{\text{pre}}$ and minimize
$$  \mathcal{L}^{(i_\star)}(\theta) = \omega_{\mathrm{PDE}}\mathcal{L}_{\mathrm{PDE}}^{(i_\star)}(\theta) + \omega_{\mathrm{IC}}\mathcal{L}_{\mathrm{IC}}^{(i_\star)}(\theta) + \omega_{\mathrm{IC}'}\,\mathbf{1}^{\,i_\star}_{\mathrm{2nd}}\, \mathcal{L}_{\mathrm{IC}'}^{(i_\star)}(\theta)  .$$

\section{Results}
\label{sec:results}

In this section, we evaluate the proposed physics-informed foundation models across the PDE families in Table~\ref{tab:pdes}. 
Section~\ref{subsec:training_scenarios} examines performance under three limited-data regimes (sparse labels, partially labeled domains, and few function pairs). 
Section~\ref{subsec:noise} assesses robustness to label noise. 
Section~\ref{subsec:resample_num_points} analyzes how resampling PDE collocation points and varying their number affect accuracy. 
Section~\ref{subsec:accuracy_efficiency} compares differentiation backends (FDM vs.\ AD) in terms of accuracy and computational cost. 
Finally, Section~\ref{subsec:downstream} studies zero-shot transfer to unseen PDE families via physics-only zero-shot fine-tuning. The code in this study will be publicly available at the GitHub repository \url{https://github.com/lu-group/pde-foundation-model}. 

To evaluate the performance of the foundation models, we adopt the metrics of $L^2$ relative error and $H^1$ relative error.
The $L^2$ relative error is a common metric used to measure the accuracy of approximate solutions to PDEs. It measures, in the sense of $L^2$ norm, how far the approximate solution $v$ deviates from the ground-truth solution $u$. The $L^2$ norm of a function $u(x)$ over a domain $\Omega$ is defined as
$$
\| u \|_{L^2} = \left( \int_{\Omega} u^2 \, dx \right)^{\frac{1}{2}}.
$$
Using this norm, the $L^2$ relative error between the exact solution $u$ and the approximate solution $v$ is given by
$$
L^2 \text{ Relative Error} = \frac{\| u - v \|_{L^2}}{\| u \|_{L^2}}.
$$
However, $L^2$ relative error does not account for errors in derivatives, making it less suitable for problems where gradient accuracy is crucial. Thus, we also adopt the $H^1$ relative error.

The $H^1$ relative error is a measure to evaluate the accuracy of an approximate solution in the Sobolev space $ H^1 $. This space accounts for both the function values and their first derivatives, making the $ H^1 $ norm a common choice for problems involving PDEs.
The $ H^1 $ norm of a function $ u $ is typically defined as
$$
\|u\|_{H^1} = \left( \|u\|_{L^2}^2 + \|\nabla u\|_{L^2}^2 \right)^{1/2}.
$$
The $H^1$ relative error quantifies the difference between an approximate solution $ v $ and the ground truth $ u $. It is expressed as
$$
H^1 \text{ Relative Error} = \frac{\|u - v\|_{H^1}}{\|u\|_{H^1}}.
$$

\subsection{Training with sparse data}
\label{subsec:training_scenarios}
The performance of purely data-driven methods typically relies heavily on the quality and quantity of available data. In this section, we construct three limited-data scenarios: sparse labeled data (Section~\ref{subsec:sparse_labeled}), partially labeled domains (Section~\ref{subsec:partially_labeled}), and limited samples in the function space (Section~\ref{subsec:small_func}). We demonstrate that in all three cases, data-driven foundation models struggle under data scarcity, whereas our physics-informed foundation models continue to perform reliably, even with minimal data.

For training the network parameters $\theta$, we employ the AdamW optimizer~\cite{loshchilov2017decoupled}. This algorithm is selected for its robust performance and its improved handling of regularization by decoupling weight decay from the adaptive learning rate mechanism.
The optimizer was configured with the following parameters: a base learning rate of $\alpha = 1 \times 10^{-4}$, a weight decay factor of $1 \times 10^{-4}$, exponential decay rates for moment estimates $\beta_1 = 0.9$ and $\beta_2 = 0.999$, and an epsilon of $\epsilon = 1 \times 10^{-6}$ for numerical stability.

To dynamically adjust the learning rate $\alpha$ throughout training, we utilize a linear warmup followed by a cosine decay schedule. The learning rate is linearly increased from an initial value to the base rate of $1 \times 10^{-4}$ over the first $T_{\text{warmup}} = 3200$ iterations. Subsequent to the warmup phase, the learning rate is annealed following a cosine curve over the remaining training steps, reaching a near-zero value at the maximum number of iterations, $T_{\text{max}} = 32000$. The learning rate $\alpha_t$ at a given iteration $t > T_{\text{warmup}}$ is given by
\begin{equation*}
\alpha_t = \frac{1}{2} \left(1 + \cos\left(\frac{t - T_{\text{warmup}}}{T_{\text{max}} - T_{\text{warmup}}}\pi\right)\right) \alpha.
\end{equation*}
This scheduling policy allows for stable convergence during the initial training phase while facilitating fine-tuning as the model approaches a minimum.

A total of 10 PDE families (Table~\ref{tab:pdes})---Adv, Diff, Diff--Lin, Diff--Log, Diff--SLog, Cons--Cub, Cons--Lin, Cons--Sin, KG, and SG---are used for training and testing of the three scenarios. 
Our dataset is defined on a uniform grid of size $64 \times 128$, corresponding to the spatiotemporal domain $[0,1] \times [0,1]$. The temporal grid points are given by
$$
\mathcal{T}_t = \{t_0, t_1, \ldots, t_{63}\} = \left\{0, \tfrac{1}{64}, \ldots, \tfrac{63}{64} \right\},
$$
and the spatial grid points by
$$
\mathcal{T}_x = \{x_0, x_1, \ldots, x_{127}\} = \left\{0, \tfrac{1}{128}, \ldots, \tfrac{127}{128} \right\}.
$$

The labeled spatiotemporal points $\mathcal{T}_{s,u_0}^{\mathrm{data}}$ for training are set to be
$$
\mathcal{T}_{\mathrm{train}} = \{t_0, t_2, \ldots, t_{62}\} \times \mathcal{T}_x 
$$
or a subset thereof, depending on data availability in each training scenario. The models are evaluated on the complementary grid
$$
\mathcal{T}_{\mathrm{test}} = \{t_1, t_3, \ldots, t_{63}\} \times \mathcal{T}_x,
$$
where $\times$ denotes the Cartesian product. Notably, training and testing points use alternating time slices: we train on even-indexed time stamps and evaluate on the complementary odd-indexed ones. This is more challenging than the standard protocol that tests on the same time stamps, as it requires generalization to unseen temporal locations. For the initial condition inputs, we choose $\mathcal{T}_{s,u_0}^{\mathrm{IC}} = \mathcal{T}_{s,u_0}^{\mathrm{IC^\prime}} = \mathcal{T}_x $ as this spatial grid is sufficiently dense and computationally efficient. Therefore, the grids $\mathcal{T}_{\mathrm{train}}$, $\mathcal{T}_{s,u_0}^{\mathrm{IC}}$, and $\mathcal{T}_{s,u_0}^{\mathrm{IC^\prime}}$ remain fixed throughout training.

For $\mathcal{T}_{s,u_0}^{\mathrm{PDE}}$ in the PDE loss term, we randomly sample 500 collocation points from the domain $[0,1] \times [0,1]$ and resamples all points at each training iteration. The effects of the resampling strategy and the number of collocation points on training performance are discussed in Section~\ref{subsec:resample_num_points}.

\subsubsection{Training with sparse labeled spatiotemporal data}
\label{subsec:sparse_labeled}

In this scenario, the labeled training data are sampled at varying spatiotemporal resolutions. Specifically, we consider uniform spatiotemporal grids of sizes $2 \times 8$, $4 \times 16$, $8 \times 32$, $16 \times 64$, and $32 \times 128$, where each pair denotes the number of temporal and spatial points, respectively. For example, the $2 \times 8$ grid corresponds to the subset of $\mathcal{T}_{\mathrm{train}}$
$$
\{t_0, t_{32}\} \times \{x_0, x_{16}, x_{32}, x_{48}, x_{64}, x_{80}, x_{96}, x_{112}\},
$$
which is uniformly spaced in both time and space.

For each PDE family, the training, validation, and testing datasets comprise 50,000, 10,000, and 10,000 labeled input-output function pairs, respectively. Across all 10 PDE families, this amounts to 500,000 training pairs, 100,000 validation pairs, and 100,000 testing pairs.
Training is conducted with a batch size of 128 over 32,000 iterations. At each iteration, the training subsets \textcolor{black}{$\mathcal{Z}^{\mathrm{PDE}} = \mathcal{Z}^{\mathrm{IC}} = \mathcal{Z}^{\mathrm{IC}^\prime} = \mathcal{Z}^{\mathrm{data}}$ }are randomly resampled, each containing 128 samples.

\begin{figure}[htbp]
    \centering
    \includegraphics[width=\textwidth]{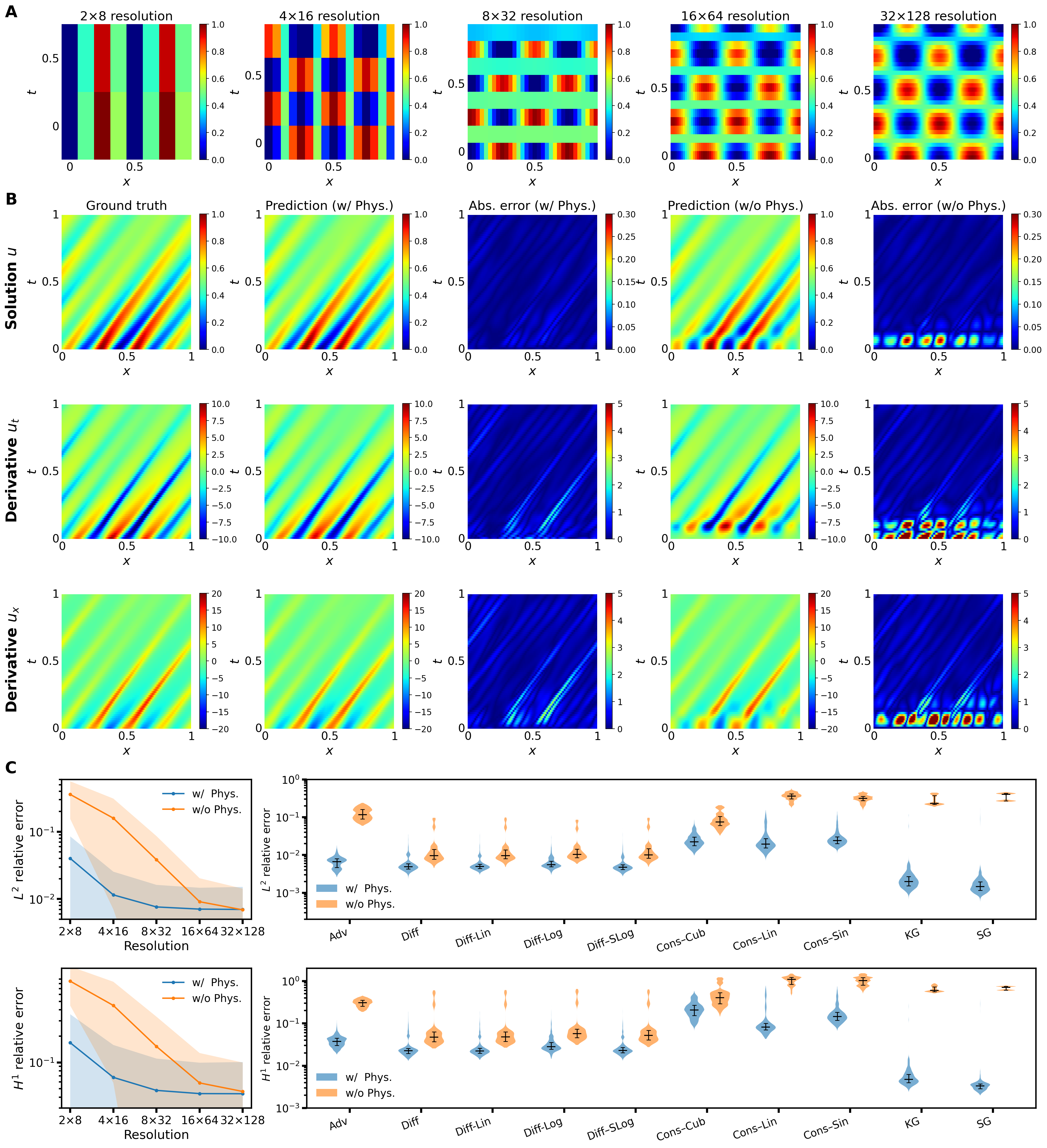}
    \caption{\textbf{Comparison of PI-MFM (w/ physics) and purely data-driven (w/o physics) models in the sparse data regime.} (\textbf{A}) An illustrative SG example under increasing resolution. (\textbf{B}) A representative Cons-Sin example from the test set, where the models are trained on labeled data of $8 \times 32$ resolution. 
    % The figures show the ground truth, prediction, and absolute error for solution $u$, temporal derivative $u_t$, and spatial derivative $u_x$. 
    The physics-informed model accurately predicts the solution and its derivatives, while the purely data-driven model fails, resulting in large errors. 
    (\textbf{C}) $L^2$ and $H^1$ relative test errors for physics-informed and purely data-driven models. (Left) At each labeled resolution, the mean test error over all 10 PDE families, with shaded bands indicating the corresponding standard deviation across PDEs. (Right) Distribution of errors for each of the 10 PDE families at $4\times16$ resolution.}
    \label{fig:sparse_data}
\end{figure}

We evaluate the model's ability to learn from limited labeled data by training it on sparse spatiotemporal grids of varying resolutions (Fig.~\ref{fig:sparse_data}). To visualize the training data sparsity, Fig.~\ref{fig:sparse_data}A shows an illustrative Sine-Gordon example under increasing resolution. At the coarsest $2 \times 8$ resolution, the data points are extremely sparse, providing minimal information about the underlying dynamics, whereas the solution's structure becomes clear at the $32 \times 128$ resolution. 

A qualitative example for the conservation law with sine flux is shown in Fig.~\ref{fig:sparse_data}B. It compares the ground truth solution $u$ and its derivatives $u_t$ and $u_x$ with predictions from physics-informed (``w/ Phys.'') and purely data-driven (``w/o Phys.'') models. The models are trained on labeled data of $8 \times 32$ spatiotemporal resolution. The physics-informed model successfully reconstructs the solution and accurately captures the complex structures of its temporal and spatial derivatives, resulting in low absolute errors. In contrast, the model trained only on data fails to generalize correctly from the sparse observations, producing predictions with significant artifacts and large errors, especially for the derivatives. 

The left of Fig.~\ref{fig:sparse_data}C illustrates the average test error as a function of the training data resolution on the 100,000 testing pairs. Both the $L^2$ and $H^1$ relative error demonstrate that incorporating physics losses leads to substantially lower errors compared to the purely data-driven model. The advantage of the physics-informed approach is particularly remarkable at lower resolutions, where the model effectively leverages the underlying physical laws to compensate for the scarcity of labeled data. The shaded regions around each curve denote the mean $\pm$ one standard deviation of the test error at each resolution, computed over all test samples from the 10 PDE families. The seemingly large standard deviations (comparable to the mean) arise because, at each resolution, both the mean and standard deviation are computed from the test errors of all 10 PDE families combined. 

We further decompose the performance by PDE family via the distribution of test errors across all 10 PDE families (Fig.~\ref{fig:sparse_data}C Right). For each case, the long horizontal bar denotes the median error, while the two shorter bars indicate the 25th and 75th percentiles. Both physics-informed and data-driven models are trained on labeled data of $4 \times 16$ spatiotemporal resolution. For every family, the physics-informed model achieves a lower median error than the data-driven baseline. In most cases, the physics-informed model also exhibits a reduced spread.

\subsubsection{Training with partially labeled temporal domain}
\label{subsec:partially_labeled}

In this section, the problem setup closely follows Section~\ref{subsec:sparse_labeled} except for the spatiotemporal grids of the labeled training data. Specifically, we select the first 2, 4, 8, 16, or 32 temporal grid points from the set $\{t_0, t_2, \ldots, t_{62}\}$, combined with the full spatial grid $\mathcal{T}_x = \{x_0, x_1, \ldots, x_{127}\}$. For instance, selecting the first 8 temporal steps results in the labeled training grid:
$$
\mathcal{T}_{\mathrm{train}} = \{t_0, t_2, t_4, t_6, t_8, t_{10}, t_{12}, t_{14} \} \times \mathcal{T}_x 
$$
These choices of 2, 4, 8, 16, and 32 temporal steps correspond respectively to the temporal intervals $[0, 1/16]$, $[0, 1/8]$, $[0, 1/4]$, $[0, 1/2]$, and $[0, 1]$.

\begin{figure}[htbp]
    \centering
    \includegraphics[width=\textwidth]{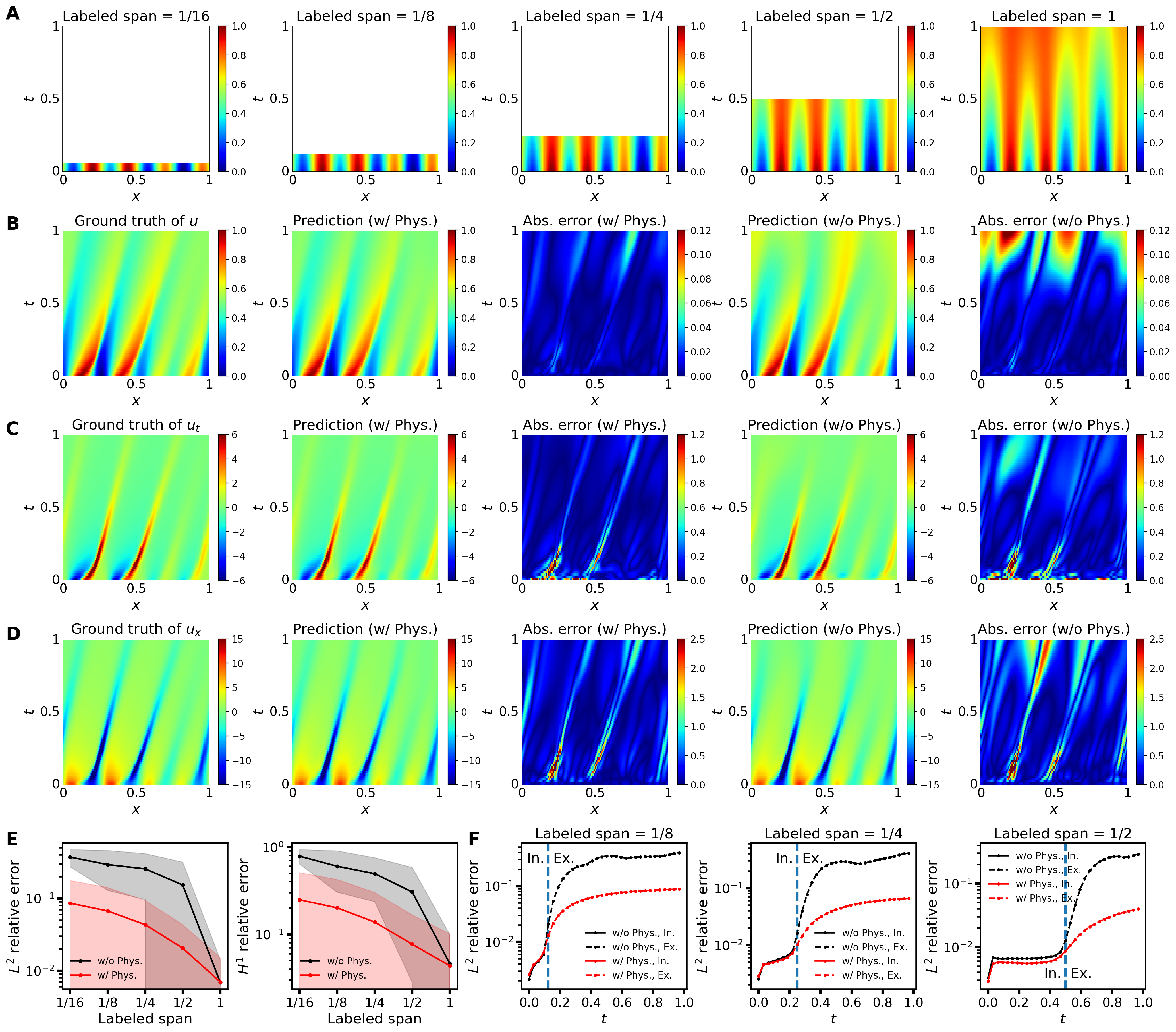}
    \caption{\textbf{Comparison of PI-MFM (w/ physics) and purely data-driven (w/o physics) models in the temporal extrapolation regime.} (\textbf{A}) An illustrative Diff-Log example under increasing labeled temporal domain. (\textbf{B}, \textbf{C}, and \textbf{D}) A representative Cons-Cub example from the test set, where the models are trained on labeled data from the temporal domain $[0, 1/2]$. (B) Solution $u$. (C) Temporal derivative $u_t$. (D) Spatial derivative $u_x$. The physics-informed model accurately extrapolates the solution and its derivatives beyond the training domain, while the purely data-driven model fails. (\textbf{E}) Mean $L^2$ and $H^1$ relative test errors over the full temporal domain versus the temporal span of labeled training data. The physics-informed model achieves significantly lower error than the purely data-driven model. (\textbf{F}) Mean $L^2$ relative error over the full temporal domain for models trained on temporal spans of 1/8, 1/4, and 1/2, respectively, highlighting the low and stable errors in both the interpolation (In.) and extrapolation (Ex.) regimes for physics-informed models.}
    \label{fig:partial_domain}
\end{figure}

Fig.~\ref{fig:partial_domain} illustrates the model's performance in temporal extrapolation, where it is trained on data from a limited initial temporal span and tested over the full domain. Fig.~\ref{fig:partial_domain}A visually depicts the partial labeled data for a sample of diffusion logistic reaction equation, showing the increasing amount of training information as the temporal span grows.
A qualitative example of this extrapolation capability is presented for the conservation law with cubic flux equation in Figs.~\ref{fig:partial_domain}B--D. Even when trained on a partial temporal domain $[0, 1/2]$, the physics-informed model accurately reconstructs the solution $u$ and its derivatives $u_t$ and $u_x$ across the entire spatiotemporal domain. The purely data-driven model, however, fails to capture the correct dynamics beyond its labeled training domain, resulting in significant prediction errors.

The overall $L^2$ and $H^1$ relative errors on the 100,000 testing pairs (Fig.~\ref{fig:partial_domain}E) decrease as the labeled temporal span for training increases. Critically, the model incorporating the PDE residual loss (``w/ Phys.'') consistently outperforms the purely data-driven model (``w/o Phys.''), with the performance gap being most significant for shorter labeled training temporal spans.
A more detailed analysis of the error evolution is provided in Fig.~\ref{fig:partial_domain}F, which shows the $L^2$ relative error over time on the test dataset for models trained on temporal spans of $1/8$, $1/4$, and $1/2$, respectively. For the physics-informed model, the error remains relatively low and stable in both the interpolation (``In.'') and extrapolation (``Ex.'') regimes. In contrast, the error for the purely data-driven model diverges rapidly immediately outside the labeled training interval.

\subsubsection{Training with sparse data in the function space}
\label{subsec:small_func}

In this scenario, the number of labeled input-output function pairs per PDE family, denoted by $N_\text{func}$, varies from 10 to 50,000. For each PDE family, the train, validation, and test datasets consist of $N_\text{func}$, 10,000, and 10,000 labeled input-output function pairs, respectively. Aggregated across all 10 PDE families, this yields a total of $10N_\text{func}$ training pairs, 100,000 validation pairs, and 100,000 testing pairs. The labeled training grid is set to be $\mathcal{T}_{\mathrm{train}} = \{t_0, t_2, \ldots, t_{62}\} \times \mathcal{T}_x$.

Training is conducted over 32,000 iterations. At each iteration, subsets \textcolor{black}{$\mathcal{Z}^{\mathrm{PDE}} = \mathcal{Z}^{\mathrm{IC}} = \mathcal{Z}^{\mathrm{IC}^\prime}$} are randomly resampled, each containing 128 samples. Due to the limited labeled training data pairs, subsets $\mathcal{Z}^{\mathrm{data}}$ are randomly resampled with a small batch size of 100.

\begin{figure}[htbp]
    \centering
    \includegraphics[width=\textwidth]{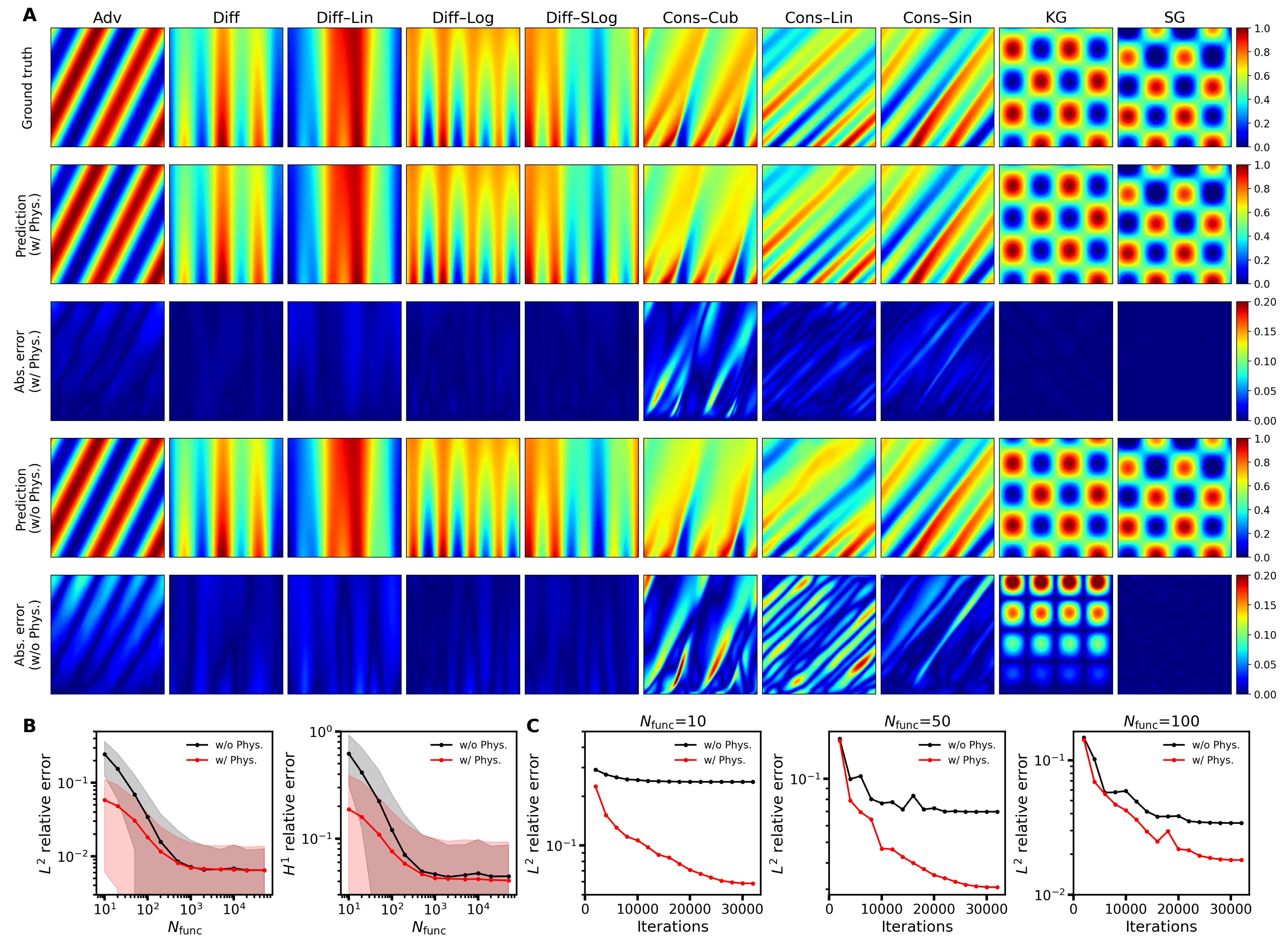}
    \caption{\textbf{Comparison of PI-MFM (w/ physics) and purely data-driven (w/o physics) models with varying numbers of labeled function pairs.} (\textbf{A}) Representative test-set examples across 10 PDE families, with models trained on 100 labeled function pairs per family. The physics-informed model provides more accurate predictions across the PDE families. (\textbf{B}) Mean $L^2$ and $H^1$ relative test errors on the test dataset versus the number of labeled input-output function pairs per PDE family $N_\text{func}$. The physics-informed model (`w/ Phys.') significantly outperforms the data-driven model (`w/o Phys.') when $N_\text{func}$ is small. (\textbf{C}) Mean $L^2$ relative error on the validation dataset versus training iterations for $N_\text{func}$ of 10, 50, and 100, respectively, showing that the physics-informed model's error continues to decrease while the data-driven model's error plateaus, particularly with fewer function pairs.}
    \label{fig:func_space}
\end{figure}

We first demonstrate the model's performance when trained with different $N_\text{func}$ by providing representative test-set examples across 10 PDE families (Fig.~\ref{fig:func_space}A). For each family, the ground truth solution is compared against predictions from models trained with and without the physics loss when $N_\text{func}=100$, along with their respective absolute errors. These visualizations confirm that the physics-informed model yields more accurate predictions across the PDE families. For instance, in the case of the Klein-Gordon (KG) family, which exhibits an oscillatory spatiotemporal pattern, the prediction from the physics-informed model closely matches the ground truth. In contrast, the prediction from the model trained without physics exhibits a noticeable phase shift along the time dimension (vertical axis) compared to the ground truth. This misalignment leads to significantly higher absolute errors, underscoring the physics-informed model's superior ability to capture both the structure and correct temporal evolution of complex dynamics.

We show the mean $L^2$ and $H^1$ relative errors on the test dataset as a function of $N_\text{func}$ (Fig.~\ref{fig:func_space}B). Both error metrics decrease as more labeled function pairs are used for training, particularly in the range of $N_\text{func}$ from 10 to approximately 2000. However, when $N_\text{func}$ exceeds roughly 2000, the rate of error reduction diminishes substantially, and the error curves begin to plateau. Notably, the model incorporating the PDE residual loss (``w/ Phys.'') consistently achieves lower errors than the purely data-driven model (``w/o Phys.'') when $N_\text{func}$ is small, highlighting the benefit of physics-informed learning in data-scarce scenarios.

The learning dynamics are further illustrated in Fig.~\ref{fig:func_space}C, which display the $L^2$ relative error on the validation dataset versus the number of training iterations for $N_\text{func}$ values of 10, 50, and 100, respectively. While the error for the model trained without physics tends to reach a steady state after a certain number of iterations, the error for the physics-informed model continues to decrease. Furthermore, the point at which the data-driven model's error plateaus occurs earlier when fewer labeled function pairs are available.

\subsection{Training with noisy data}
\label{subsec:noise}

\begin{figure}[htbp]
    \centering
    \includegraphics[width=\textwidth]{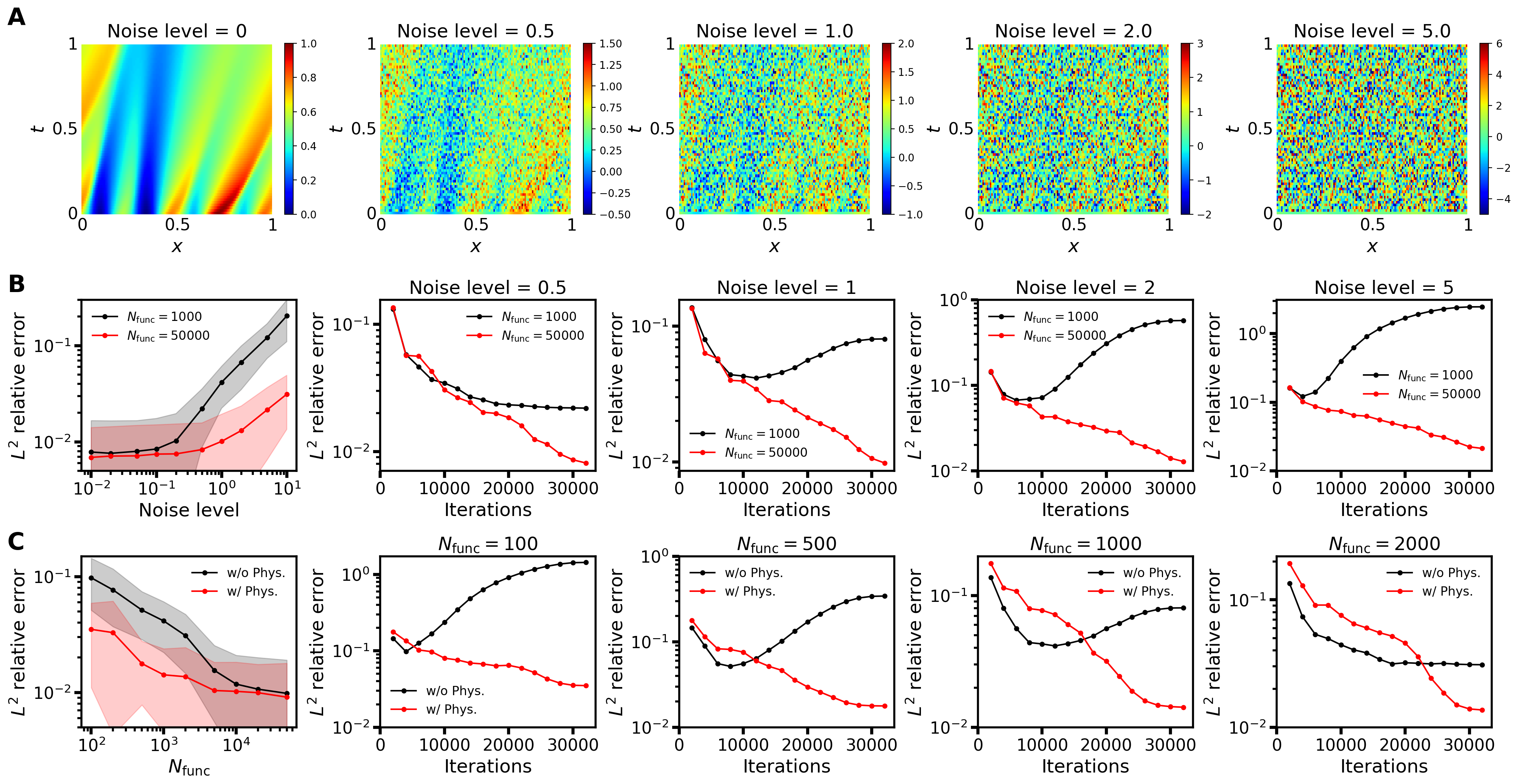}
    \caption{\textbf{Robustness to data noise.} (\textbf{A}) An illustrative  Cons-Cub example with increasing label noise level $\gamma$ from 0 to 5. (\textbf{B}) The first figure reports mean $L^{2}$ relative test error versus $\gamma$ for $N_{\text{func}}=1000$ and $50000$. The remaining four figures show training curves at $\gamma\in\{0.5,1,2,5\}$, where larger $N_{\text{func}}$ avoids overfitting. (\textbf{C}) Fixing $\gamma=1$. The first figure shows mean $L^{2}$ relative test error versus $N_{\text{func}}$. The remaining four figures compare data-driven (`w/o Phys.') and physics-informed (`w/ Phys.') training curves at $N_{\text{func}}\in\{100,500,1000,2000\}$, highlighting the stability of physics-informed training.}
    \label{fig:noise}
\end{figure}

To test robustness, we corrupt the function observations used in the data loss $\mathcal{L}_{\mathrm{data}}$ with additive white Gaussian noise whose magnitude is controlled by a dimensionless noise level $\gamma$. Specifically, let $D\in\mathbb{R}^m$ denote the clean observations, where $m$ is the total number of observed entries. We draw i.i.d. standard normal noise $z\sim\mathcal{N}(0,I_m)$ and rescale it so that its $L^2$ norm is a fixed fraction of the signal norm:
$$
\tilde D \;=\; D \;+\; \gamma\,\frac{\lVert D\rVert_2}{\lVert z\rVert_2}\,z,
\qquad\text{so that}\qquad
\frac{\lVert \tilde D-D\rVert_2}{\lVert D\rVert_2}=\gamma .
$$
Thus, the noise level $\gamma$ means the $L^2$ relative noise magnitude. Equivalently, the signal-to-noise ratio is $\mathrm{SNR}=1/\gamma$. The scaling makes the total noise energy proportional to the signal energy and independent of grid resolution. We use $\gamma\in[10^{-2},10^{1}]$ on a logarithmic grid and include $\gamma=0$ as a clean baseline. Noise is applied only to the supervised function values; the physics residual points and hard boundary/initial conditions are left noise-free. Fig.~\ref{fig:noise}A shows a representative Cons-Cub example at increasing noise levels. At $\gamma=0$ the field exhibits smooth, coherent spatiotemporal structure. At $\gamma$ = 0.5 and 1, speckled perturbations are visible but large-scale patterns are still discernible. At $\gamma$ = 2 and $5$, the labels are dominated by high-variance fluctuations that largely obscure fine features.

In this scenario, $N_\text{func}$ varies from 100 to 50,000. For each PDE family, the training, validation, and test datasets consist of $N_\text{func}$, 10,000, and 10,000 labeled input-output function pairs, respectively. Aggregated across all 10 PDE families, this yields a total of $10N_\text{func}$ training pairs, 100,000 validation pairs, and 100,000 testing pairs. The labeled training grid is set to be $\mathcal{T}_{\mathrm{train}} = \{t_0, t_2, \ldots, t_{62}\} \times \mathcal{T}_x$. Training is conducted with a batch size of 128 over 32,000 iterations. At each iteration, the training subsets $\mathcal{Z}^{\mathrm{PDE}} = \mathcal{Z}^{\mathrm{IC}} = \mathcal{Z}^{\mathrm{IC}^\prime} = \mathcal{Z}^{\mathrm{data}}$ are randomly resampled, each containing 128 samples.

We first investigate the effect of noise on generalization and training dynamics (Fig.~\ref{fig:noise}B). Errors rise with the noise level $\gamma$ but abundant data markedly improves robustness. For the network training at $\gamma\in\{0.5,1,2,5\}$, with $N_{\text{func}}=1000$, the validation error typically decreases and then rebounds, indicating overfitting to the noisy labels. With $N_{\text{func}}=50000$, the error continues to descend and no overfitting is observed. Even at $\gamma=10$ (i.e., 1000\% label noise), the model trained with $N_{\text{func}}=50000$ attains about 3\% test $L^2$ error.

We then fix $\gamma=1$ and varies $N_{\text{func}}$ (Fig.~\ref{fig:noise}C). Increasing $N_{\text{func}}$ consistently reduces test error, with diminishing returns beyond a few thousand samples. We also compare training curves for the data-driven model (w/o Phys.) and the physics-informed model (w/ Phys.) at $N_{\text{func}}\in\{100,500,1000,2000\}$. For small to moderate $N_{\text{func}}$ ($=100,500,1000$), the data-driven model overfits (error bottoms out and then rises), whereas the physics-informed model remains stable and continues to improve. These results indicate that the PDE residual acts as an effective physics-guided prior and regularizer under label noise, especially when labeled data are limited.

\subsection{Effectiveness of PDE collocation point resampling}
\label{subsec:resample_num_points}

In Section~\ref{subsec:workflow_algorithm}, we emphasized that the choice of $\mathcal{T}_{s,u_0}^{\mathrm{PDE}}$ is highly flexible and can significantly influence training performance. Here, we consider two strategies: ``Resample'' and ``Non-resample''. In the ``Resample'' setting, $\mathcal{T}_{s,u_0}^{\mathrm{PDE}}$ is randomly resampled at each training iteration, while in ``Non-resample'' setting, $\mathcal{T}_{s,u_0}^{\mathrm{PDE}}$ is fixed after its initial selection and remains unchanged throughout training.

We investigate the effects of resampling strategy and the number of collocation points, $|\mathcal{T}_{s,u_0}^{\mathrm{PDE}}|$. Specifically, 
we use a uniform spatiotemporal grids of size $4 \times 16$ as described in Section~\ref{subsec:sparse_labeled}, corresponding to the subset of $\mathcal{T}_{\mathrm{train}}$:
$$\{t_0, t_{16}, t_{32}, t_{48}\} \times \{x_0, x_{8}, x_{16}, x_{24}, \ldots, x_{96}, x_{104}, x_{112}, x_{120}\},$$ 
which is evenly spaced in both time and space. 

\begin{figure}[htbp]
    \centering
    \includegraphics[width=\textwidth]{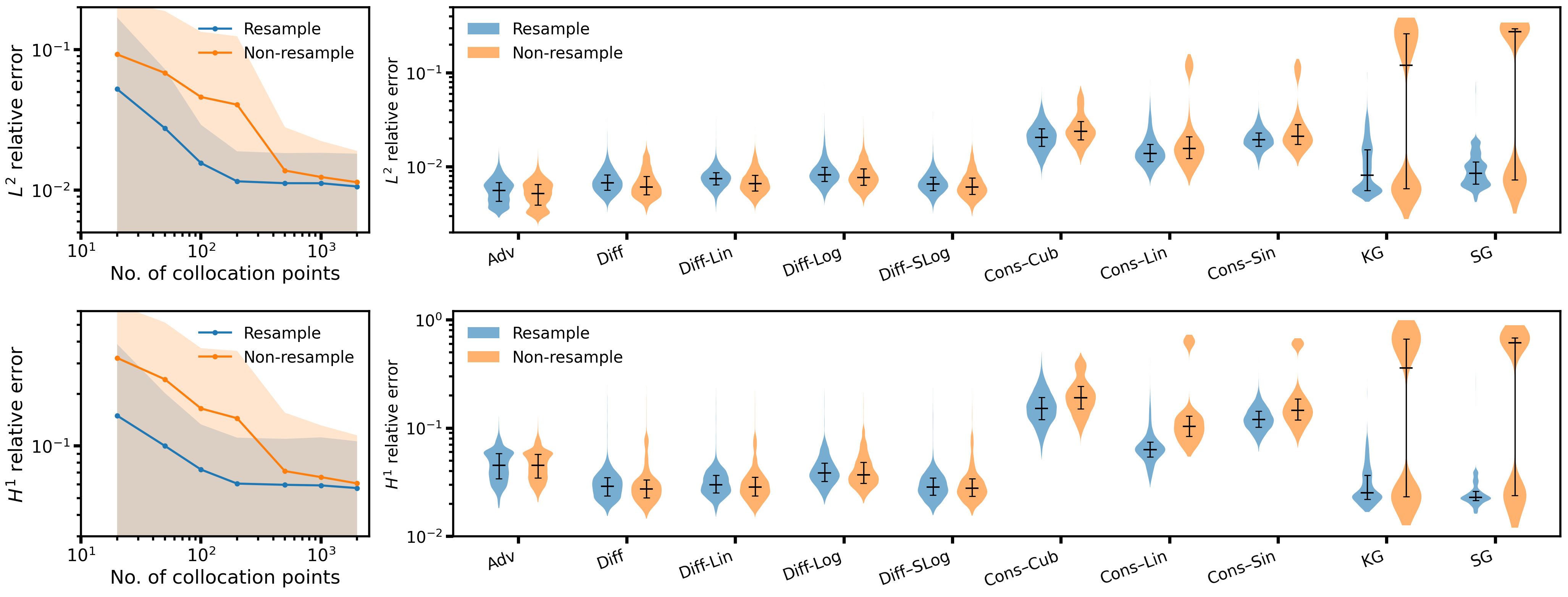}
    \caption{\textbf{Effect of collocation-point resampling and collocation-set size on test errors.} (\textbf{Left}) Mean and one standard deviation of $L^2$ (top) and $H^1$ (bottom) relative test errors versus the number of PDE collocation points, comparing iteration-wise ``Resample'' against fixed ``Non-resample''. Increasing the number of collocation points reduces errors for both strategies, while resampling consistently achieves lower errors at a given collocation budget. (\textbf{Right}) Distributions of test errors across the 10 PDE families under the two strategies, illustrating that resampling improves both accuracy and robustness across families and suppresses outliers observed with fixed collocation sets.}
    \label{fig:resample_num_points}
\end{figure}

We first present the impact of the collocation point sampling strategy on the model's test performance (Fig.~\ref{fig:resample_num_points} Left). The number of collocation points $|\mathcal{T}_{s,u_0}^{\mathrm{PDE}}|$ varies from 20 to 2000. The results for $L^2$ and $H^1$ relative test errors demonstrate a trend that increasing the number of collocation points leads to a reduction in test error for both strategies. However, the ``Resample'' strategy consistently and significantly outperforms the ``Non-resample'' approach. For any given number of collocation points, randomly resampling these points at each step of training yields a substantially more accurate solution. This suggests that dynamically exploring the spatiotemporal domain to enforce the PDE residual is a more effective and efficient training methodology than relying on a static set of points.

We also provide a family-wise view of the test error distributions for each of the 10 PDE families (Fig.~\ref{fig:resample_num_points} Right). Overall, resampling shifts the error distributions downward and, importantly, reduces their spread across most families, indicating improved robustness. The effect is particularly evident in the heavier-tailed families (e.g., KG and SG), where a fixed collocation set can lead to large-error outliers that are substantially mitigated by iteration-wise resampling.

\subsection{Accuracy and efficiency of differentiation methods}
\label{subsec:accuracy_efficiency}

We now examine how the differentiation method influences accuracy and computational efficiency. Following Section~\ref{subsec:differentiation_methods}, we compare finite-difference approximations (FDM) computed in float64, float32, float16, and bfloat16 against automatic differentiation (AD). We use the dataset and training settings of Section~\ref{subsec:sparse_labeled}, with a $4\times16$ spatiotemporal grid for labeled data.

\begin{figure}[htbp]
    \centering
    \includegraphics[width=.8\textwidth]{figs/step_size.png}
    \caption{\textbf{Effects of FDM step sizes for accuracy.} Mean $L^2$ and $H^1$ relative test errors versus the finite-difference step size for float64, float32, float16, and bfloat16. The dashed line is the step-size-free AD baseline. All FDM curves are U-shaped: for large step sizes, the error is truncation-dominated, while for small step size, the error is round-off-dominated. The optimal step size depends on precision: float64 achieves a broad minimum that nearly matches AD, and float32 attains similar best accuracy with a comparably broad range, whereas float16/bfloat16 exhibit much narrower safe step-size windows and higher error floors, especially for $H^1$.}
    \label{fig:step_size}
\end{figure}

We summarize the mean $L^2$ and $H^1$ relative test errors for different FDM step size $\Delta x=\Delta t=\Delta$ (Fig.~\ref{fig:step_size}). Each FDM curve exhibits the expected U-shape: when $\Delta$ is large, truncation dominates and the error is high; as $\Delta$ decreases, the error drops until round-off becomes dominant, after which the curves turn upward. The location of the minimum depends strongly on precision. In float64 the minimum spans a wide range of small steps and nearly coincides with the AD baseline; in float32 the minimum shifts to slightly larger steps with nearly the same best accuracy; in float16 and bfloat16 the minimum occurs at still larger steps, and the attainable error floor is higher, especially for $H^1$. Hence, a properly tuned float32/float64 FDM approaches the AD baseline, whereas half-precision FDM has a narrower “safe” window and cannot match AD at the smallest errors. These trends are consistent with the error analysis in Section~\ref{subsec:fdm}. We list the required range for the step size in Table~\ref{tab:efficiency_h200} (``step size range'' column).

\begin{table}[htbp]
\small
\centering
\caption{\textbf{Accuracy and efficiency comparison of FDM and AD on an NVIDIA H200 SXM GPU with 141 GiB GPU memory.} We report GPU memory usage, training time per iteration, and relative errors in $L^2$ and $H^1$. For FDM, the step size is tuned and we report the range of values that achieves the best accuracy, while AD requires no such tuning.}
\label{tab:efficiency_h200}
\begin{tabular}{c|ccccc}
\toprule
\multirow{2}{*}{Train method} & GPU memory & Train time & \multirow{2}{*}{$L^2$ rel err} & \multirow{2}{*}{$H^1$ rel err} & \multirow{2}{*}{Step size range}\\
 &  (GiB) &  (sec/iter) &  &  & \\  
\midrule
FDM bfloat16 & 10.77 & 0.094 & 3.041\% & 14.91\% & [0.06, 0.1] \\
FDM  float16 & 10.77 & 0.094 & 1.629\% & 9.491\% & [0.03, 0.06] \\
FDM  float32 & 17.96 & 0.203 & 1.041\% & 6.108\% & [0.001, 0.01] \\
FDM  float64 & 34.78 & 0.258 & 1.032\% & 5.965\% & [$10^{-7}$, 0.01] \\
\midrule
Forward AD bfloat16 & 37.45 & 0.440 & 1.068\% & 5.996\% & - \\
Forward AD  float16 & 37.45 & 0.367 & 1.024\% & 6.037\% & - \\
Forward AD  float32 & 53.60 & 0.561 & 1.040\% & 5.859\% & - \\
\midrule
% Reverse AD bfloat16 &  &  &  &  & - \\
% Reverse AD  float16 &  &  &  &  & - \\
Reverse AD  float32 & 87.34 & 0.642 & 1.072\% & 6.211\% & - \\
\bottomrule
\end{tabular}
\end{table}

We further compare FDM and AD in Table~\ref{tab:efficiency_h200}. Finite differences are more economical in time and memory. FDM with float32 uses less time and memory compared with AD, and FDM with float64 is still more cost-efficient. Their accuracy, however, depends on precision. With float32/float64, FDM reaches essentially the same $L^2$ and $H^1$ errors as AD, but in half precision the best attainable error is noticeably worse, especially in $H^1$.

Forward-mode AD delivers high accuracy independently of numeric precision: the errors are nearly unchanged from bfloat16 through float32. The price is higher resource use, including more memory and longer training time than FDM, but still substantially cheaper than reverse-mode AD. Reverse-mode AD in float32 matches forward-mode accuracy, yet remains the most expensive configuration in both memory and time even under vectorized Jacobian extraction (implemented via PyTorch \texttt{vmap}). This is consistent with the analysis in Section~\ref{subsec:ad}: to recover coordinate derivatives for all $B\times M$ outputs using reverse mode, one must evaluate VJPs for $BM$ output directions, so the cost scales with the number of outputs and cannot be eliminated by vectorization.

In short, FDM with float32 can match AD accuracy at much lower cost, but it requires step-size tuning. AD does not require hyperparameter tuning and is robust to precision, with forward mode (especially with float16) markedly more efficient than reverse mode. Therefore, we suggest either FDM with float32 (and proper step size) or forward AD with float16.

% For Korteweg-De Vries equation (third-order derivative PDE) and Cahn-Hilliard equation (fourth-order derivative PDE), float32 can still achieve good accuracy at $\Delta x_{\text{optimal}}$. But the $\Delta x_{\text{optimal}}$ increases compared to first- and second-order derivative PDEs.

\subsection{Downstream task: Solving unseen PDE families by zero-shot physics-informed fine-tuning}
\label{subsec:downstream}

In this section, we perform zero-shot fine-tuning following the setup in Section~\ref{subsec:downstream_method}. We use as initialization the physics-informed model from Section~\ref{subsec:small_func} with $N_\text{func}=100$ , and adapt it, without any labeled input-output pairs, to each unseen PDE family (Burgers, Wave, Diff-Bi) separately using only the PDE, IC, and IC$^\prime$ losses.
For each PDE family, the validation and testing datasets comprise 10,000 and 10,000 labeled input-output function pairs, respectively; the training dataset comprises 50,000 unlabeled inputs only. 
Training is conducted with a batch size of 128 over 3200 iterations. At each iteration, the training subsets \textcolor{black}{$\mathcal{Z}^{\mathrm{PDE}} = \mathcal{Z}^{\mathrm{IC}} = \mathcal{Z}^{\mathrm{IC}^\prime}$} are randomly resampled, each containing 128 samples.

\begin{figure}[htbp]
    \centering
    \includegraphics[width=\textwidth]{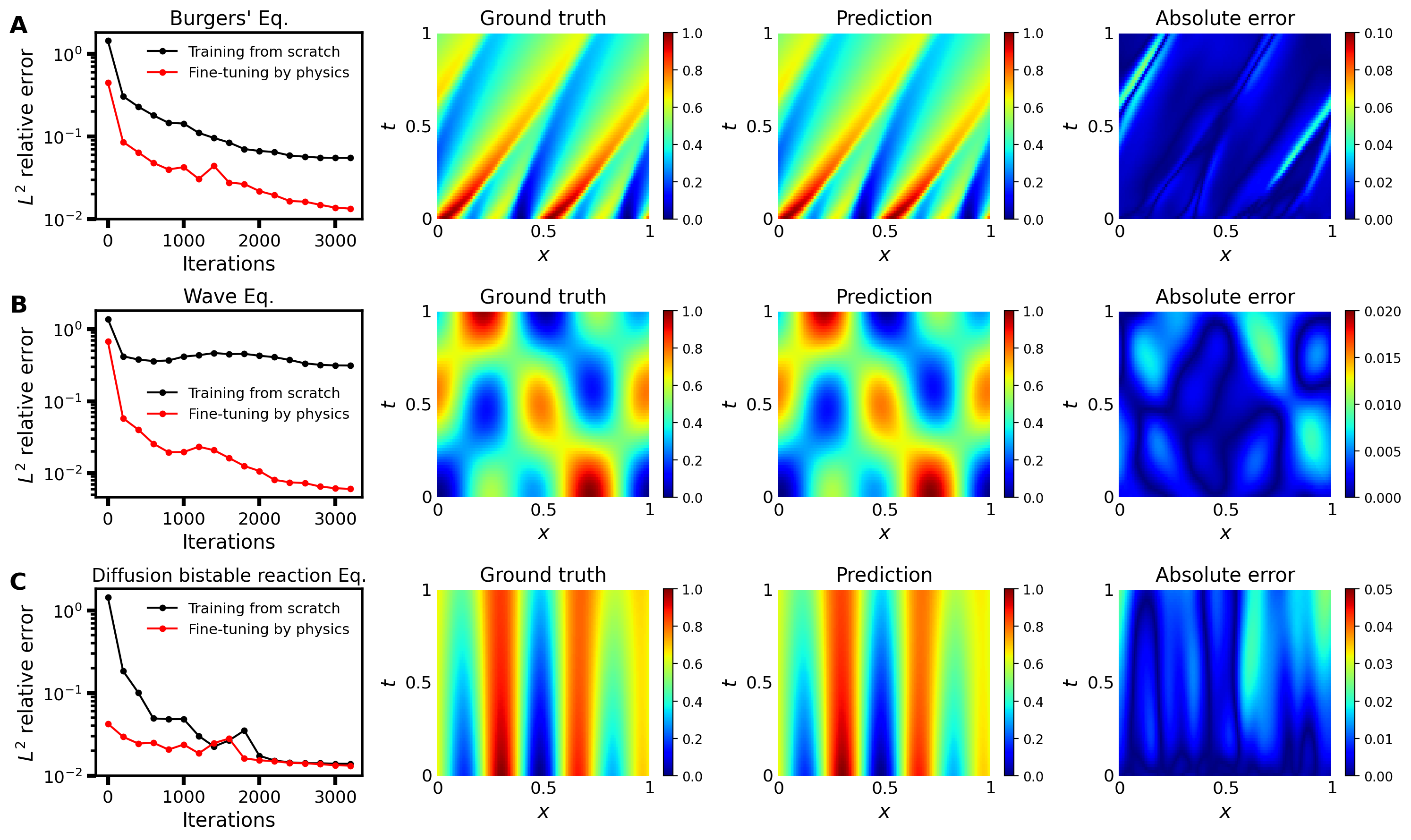}
    \caption{\textbf{Performance of physics-informed zero-shot fine-tuning on unseen PDE families.} We initialize from the physics-informed model trained with $N_{\text{func}}=100$ and adapt to each family using only the PDE and IC losses without any labeled data. (\textbf{A}) Burgers' equation. (\textbf{B}) Wave equation. (\textbf{C}) Diffusion bistable reaction equation. The first column shows mean $L^{2}$ relative test errors over 3200 iterations for zero-shot physics-informed fine-tuning versus a physics-informed baseline trained from scratch. The remaining three columns display representative test instances of zero-shot fine-tuning: ground truth, the fine-tuned prediction, and the corresponding absolute error.}

    \label{fig:downstream}
\end{figure}

We compare the test $L^{2}$ relative error of our physics-informed fine-tuning with a physics-only baseline trained from random initialization (Fig.~\ref{fig:downstream}, first column). Across all three families, our fine-tuning decreases the error rapidly (by more than an order of magnitude in the early iterations) and reaches about $1\%$ $L^2$ relative error, whereas the training from-scratch baseline converges slowly and plateaus at several percent. The points at iteration $=0$ correspond to zero-shot prediction without any adaptation. These initial errors are large, and for the Wave family they are close to $100\%$. Despite such poor starting points, the physics-informed fine-tuning reduces the error quickly to below $1\%$. We also show a representative test instance: ground truth, the model prediction after fine-tuning, and the corresponding absolute error field (Fig.~\ref{fig:downstream}). The predictions closely match the targets in all three families, and the residual errors are small and spatially localized.

% Results: Fig.~\ref{fig:downstream}

% Fine-tune the whole/part of NNs. (Using LoRA? maybe later)
% \begin{itemize}
%     \item Transfer from multiple operator learning to single operator learning with zero-shot/few-shot data
%     \item Transfer from multiple operator learning to single function approximation with zero-shot data

% \end{itemize}

% \subsubsection{Transfer to single operator learning}
% Burgers' equations (unseen in training datasets): Works with 10 known PDE cases (few-shot data) or even 0 known PDE cases (zero-shot data).

% Baseline: PINN using the same NN with random initialization.

% \subsubsection{Transfer to single function approximation with zero-shot data}
% Burgers' equations (unseen in training datasets): $L^2$ relative error $<$ 1\% within 1000 iterations.

% Baseline: Physics-informed single operator learning using the same NN with random initialization.

% \subsection{Unseen types of PDEs and transfer learning}
%Diffusion reaction equation can be extrapolated well without transfer learning. Burgers' and wave equations can be extrapolated well with transfer learning. Burgers' equation is more difficult than wave equation.

% Diff–Bi: Initial: .  After 1000 iterations of zero-shot fine-tuning: . After 3200 iterations of zero-shot fine-tuning: .

% Wave: Initial: .  After 1000 iterations of zero-shot fine-tuning: . After 3200 iterations of zero-shot fine-tuning: .

% Burgers': Initial: .  After 1000 iterations of zero-shot fine-tuning: . After 3200 iterations of zero-shot fine-tuning: .

\section{Conclusions}
\label{sec:conclusions}

We presented a physics-informed multi-operator learning framework by leveraging the governing equations. By taking a symbolic representation of the PDE as input and automatically assembling the corresponding residual, the approach avoids hand-crafted, equation-specific objectives and scales across PDE families. Although this study uses PROSE as the network architecture, the framework does not depend on the network architecture: any multimodal foundation model that accepts operator information can be trained or adapted in the same way without architectural changes.

Across sparse spatiotemporal grids, partially labeled temporal domains, and limited-function regimes, the physics-informed variants consistently and substantially outperform purely data-driven counterparts under matched budgets.
Beyond accuracy, the physics-informed objective also stabilizes optimization in the presence of substantial label noise. We further observe that simple protocol choices, such as periodically resampling collocation points, reliably tighten PDE satisfaction without altering the underlying network. Finally, our study of differentiation backends, together with a vectorized PDE loss pipeline that evaluates all required derivatives in a single pass, offers a practical recipe for deploying PDE residual supervision in PDE-encoding foundation models and clarifies when forward-mode automatic differentiation or carefully tuned finite differences are preferable in practice.

Another contribution is the demonstration of zero-shot fine-tuning for unseen PDE families in an explicit PDE-encoding multimodal foundation model. Starting from a physics-informed initialization and using only the PDE residual and IC/BC information, the method adapts rapidly and attains a small relative error, clearly surpassing training-from-scratch with the same physics-only supervision.

As future work, we will further improve physics-informed training by coupling PI-MFM with advanced training protocols. For example, we will develop adaptive sampling strategies over both the distribution of input PDE instances and the spatiotemporal collocation sets~\cite{lu2021deepxde,ouyang2025rams}, allocating the sampling budget toward regions with large residuals or high uncertainty to reduce cost without sacrificing accuracy. In addition, we will investigate training loss enhancements such as gradient-enhanced PINNs~\cite{yu2022gradient}, which incorporate derivative information of the PDE residual to accelerate convergence and improve solution fidelity. Motivated by recent loss landscape analyses of PINN optimization~\cite{rathore2024challenges}, we will also explore better optimizers that alleviate the ill-conditioning of the PINN loss, improving training stability and robustness.

Beyond training protocols, we will scale PI-MFM to two- and three-dimensional, time-dependent PDEs. We will also study more reliable deployment of PDE-encoding foundation models by equipping PI-MFM with reliable extrapolation~\cite{zhu2023reliable} and uncertainty quantification~\cite{winovich2025active,moya2025conformalized}. Since training large foundation models can be prohibitively expensive for a single research group, we will further explore federated scientific machine learning to enable collaborative training across distributed, non-i.i.d. local datasets without sharing raw data~\cite{zhang2025federated}. Finally, we will integrate PI-MFM with classical numerical solvers to mitigate the accuracy limitation of machine learning models, such as the neural-operator element method~\cite{ouyang2025neural}.

\section*{Acknowledgments}

This work was supported by the U.S. Department of Energy Office of Advanced Scientific Computing Research under Grants No.~DE-SC0025593 and No.~DE-SC0025592, and the U.S. National Science Foundation under Grants No.~DMS-2347833 and No.~DMS-2527294.
We thank Zhongyi Jiang for valuable discussions.

\bibliographystyle{unsrt}
\bibliography{main}

\end{document}